\documentclass[11pt]{roboparty-report}
\input{robohome-macros}
\input{robohome-titlepage}
\reporttitle{RoboHome: Whole-Home Simulation Scene and Data Generation for
  Embodied Long-Horizon Loco-Manipulation}
\reportshorttitle{ROBOHOME}
\newcommand{\rhlead}{\texorpdfstring{\textsuperscript{\dag}}{}}
\newcommand{\rhcore}{\texorpdfstring{\textsuperscript{\ddag}}{}}
\newcommand{\rhcorr}{\texorpdfstring{\textsuperscript{*}}{}}
\reportauthors{Haifeng Zhang\rhlead, Yuchen Wang\rhcore, Yuchen Bian\rhcore,
  Ningze Qi\rhcore, Shengze Xu\rhcore, Qizhao Geng\rhcore, Haoyang Luo, Hanbing Zhang,
  Qimai Yan, Zhexiong Wang, Zengxiang Huang, Rui Song, Jianjie Chen,
  Guangxin Yang, Junhan Sun, Qiaozhi Wang\rhcorr}
\reportaffiliations{\textsuperscript{\dag}Project Leader\qquad
  \textsuperscript{\ddag}Core contributors\qquad
  \textsuperscript{*}Corresponding author}
\reportcontact{\href{mailto:wangqiaozhi@roboparty.com}{\nolinkurl{wangqiaozhi@roboparty.com}}}
\reportgithub{https://github.com/Roboparty/RoboHome}

\reportlogo{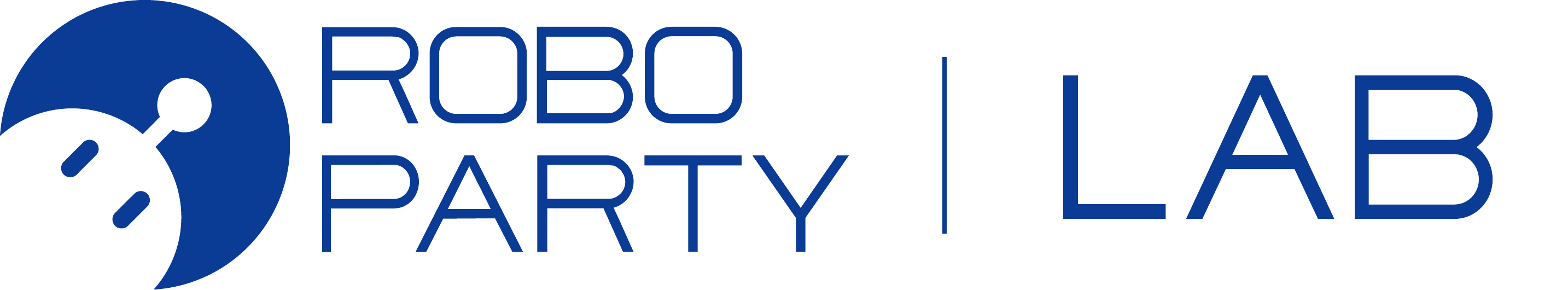}
\reportwordmark{assets/brand/roboparty-lab.pdf}
\reportheadertext{ROBOPARTY LAB}

\begin{document}
\makereporttitle
\noindent\begin{minipage}{\linewidth}
\captionsetup{type=figure}
\centering
\includegraphics[width=0.95\linewidth]{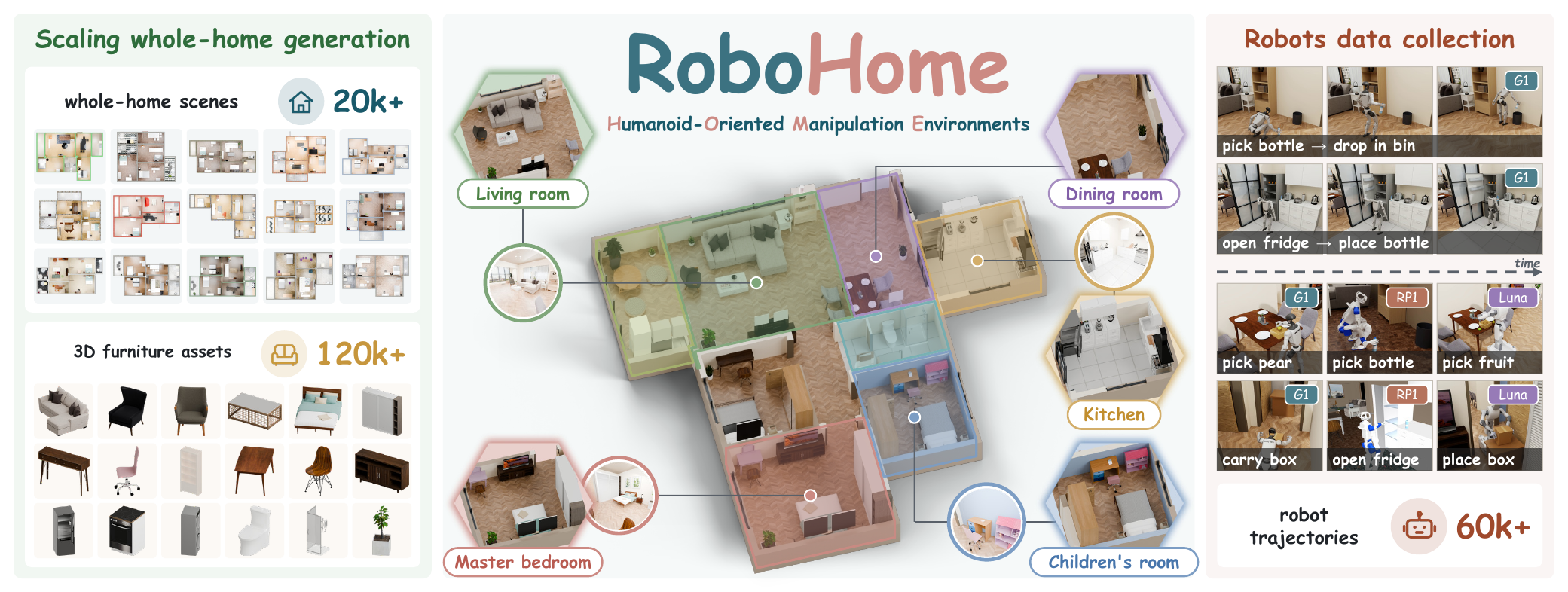}
\caption{\textbf{RoboHome overview.} \emph{Left:} whole-home scenes are generated at scale from real floorplans and original-scale 3D assets. \emph{Center:} a generated home whose rooms are interactive and traversable by the robot from the entrance. \emph{Right:} long-horizon loco-manipulation trajectories generated in these scenes for the Unitree G1, the RoboParty RP1, and the wheeled dual-arm Luna.}
\label{fig:teaser}
\end{minipage}
\par\vspace{3mm}

\begin{reportabstract}
Embodied AI is moving toward long-horizon loco-manipulation and large-scale pretraining, yet such data are costly to collect on real robots, which calls for generating whole-home scenes and trajectories automatically in simulation. We present \emph{RoboHome}, a whole-home simulation scene and data generation framework for embodied long-horizon loco-manipulation. Scene generation is constrained by robot loco-manipulation affordance: on real floorplans, it performs floorplan selection, room pre-layout, and joint asset--pose search, keeping assets at their original dimensions and checking robot traversal from the entrance. For data generation, a coding agent specifies the task goals and writes each task as a program, with generated videos providing human-like motion references; simulation ground truth, state rollback, and deterministic replay yield dynamically feasible long-horizon trajectories without human demonstrations. Over 60 generation attempts on 20 real floorplans, RoboHome saves a scene in 91.7\% of the attempts, 28.3 and 43.3 percentage points more than Holodeck and EmbodiedGen V2; its functional completeness and layout plausibility exceed the strongest baseline by 23.9 and 22.8 points, with a collision object rate of only 0.2\%. We generate data for six long-horizon tasks on the Unitree G1, the RoboParty RP1, and the wheeled dual-arm Luna. On G1, the average generation pass rate is 87.3\%, authoring the expert programs costs about \$70 of API fees per hour of generated data, and the generated trajectories are validated effective across multiple policy models.
\end{reportabstract}

\clearpage

\section{Introduction}
\label{sec:intro}

Embodied AI is moving from short-horizon tabletop manipulation~\cite{RoboTwin2,LIBERO} toward long-horizon loco-manipulation that combines locomotion with articulated-object manipulation, and increasingly relies on large-scale pretraining on simulation data~\cite{GraspVLA,InternData-A1}. Collecting such data on real robots is expensive: scene arrangements are hard to vary in bulk, and a long-horizon trajectory that fails midway must be recollected~\cite{SceneSmith}. Generating data at scale in simulation is therefore more practical, but it requires physically valid whole-home scenes and a data pipeline that efficiently produces dynamically feasible long-horizon trajectories.

Whole-home simulation requires realistic floorplans, original-scale furniture, robot accessibility, and articulated interaction. Existing resources often trade scale for realism or interaction: scans are mostly static, manually built interactive homes are limited, and generators commonly use synthetic floorplans or rescaled assets~\cite{ProcTHOR,HSSD,MolmoSpaces,BEHAVIOR-1K,HomeWorld,SceneSmith,HM3D}. Simulation data generation also depends on human demonstrations or per-skill policies~\cite{RoboTwin2,MimicGen,RoboCasa365,ManiSkill-HAB,AgentWorld}. Automatic loco-manipulation generation must couple locomotion with object interaction, compensate for whole-body tracking errors, and preserve human-like motion.

We propose RoboHome, a joint framework for whole-home scene and long-horizon loco-manipulation data generation (\cref{fig:teaser}). Its scene generation is constrained by affordance, and its data generation combines a coding agent, video motion references, and simulation feedback. Our main contributions are:
\begin{itemize}[leftmargin=*,itemsep=1pt,topsep=3pt]
\item \textbf{Affordance-constrained whole-home scene generation.} We treat robot loco-manipulation affordance as a constraint on scene generation and propose semantic-envelope pre-layout and joint asset--pose search. The method places original-scale assets in fixed normalized floorplans derived from real plans~\cite{RPLAN}, verifies entrance connectivity with the robot embodiment, and reserves clearance for opening doors and drawers.
\item \textbf{A human-like long-horizon data generation pipeline without human demonstrations.} A coding agent specifies the task goals and a demonstration prompt, and a video generation model provides human-like motion references and segment ordering. The coding agent writes tasks as programs grounded in simulation ground truth, and state rollback and deterministic replay correct failures and validate trajectories. The pipeline automatically generates dynamically feasible long-horizon loco-manipulation trajectories for humanoid and wheeled dual-arm robots.
\item \textbf{A whole-home scene asset library and a multi-embodiment long-horizon dataset.} We build a library of interactive whole-home scenes from real floorplans, together with a long-horizon loco-manipulation dataset covering three embodiments: Unitree G1, RoboParty RP1, and Luna. Experiments show that RoboHome outperforms existing methods in scene quality and saved-scene yield under the tested budget, and that the generated data support fine-tuning ACT, $\pi_{0.5}$, and GR00T N1.7 in simulation on Unitree G1.
\end{itemize}

\section{Related Work}
\label{sec:related}

\noindent\textbf{Embodied scenes and scene generation.}
Scanned environments preserve real geometry but offer limited object interaction~\cite{Matterport3D,ScanNet,HM3D,Gibson}; manually assembled interactive scenes are costly to scale~\cite{HSSD,BEHAVIOR-1K,LIVIN}. Procedural, learned, and LLM-guided methods scale generation~\cite{ProcTHOR,Infinigen-Indoors,ATISS,PhyScene,Holodeck,SceneSmith,EmbodiedGen-V2}, but often use synthetic floorplans, single rooms, or rescaled assets. Traversal-aware methods reserve walkable space or check connectivity~\cite{PhyScene,D3D-GEN,SceneSmith}. RoboHome combines real floorplans, original-scale assets, entrance connectivity, and interaction clearances (\cref{tab:scene_compare}).

\noindent\textbf{Robot simulation data generation.}
Teleoperation requires human effort~\cite{LIBERO,BEHAVIOR-1K,AgentWorld}; demonstration augmentation retains dependence on seed demonstrations~\cite{MimicGen,MoMaGen,HumanoidMimicGen,ZeroHLM}. Automatic generation uses programs or policies~\cite{GenSim2,RoboGen,RoboTwin2,InternData-A1,ManiSkill-HAB}, often for tabletop tasks or with per-skill policies. RoboHome uses agent-written programs, video references, and simulation verification to generate long-horizon loco-manipulation trajectories without human demonstrations (\cref{tab:data_compare}).

\begin{table*}[t]
\centering
\caption{Comparison of recent embodied simulation scene resources and scene generation methods (2025--2026). \cmark: supported; \xmark: not supported; \pmark: partially supported. ``Orig.\ scale'' means assets are not rescaled to model-estimated sizes; ``Emb.\ trav.'' means doorways and entrance connectivity are checked with the robot's dimensions.}
\label{tab:scene_compare}
\setlength{\tabcolsep}{4pt}
\resizebox{\textwidth}{!}{%
\begin{tabular}{@{}llcccccl@{}}
\toprule
\multicolumn{8}{c}{\textbf{Part A: Recent released scene datasets}} \\
\midrule
Resource / Method & Floorplan source & Scope & Sim-ready & Interactive art. & Orig.\ scale & Emb.\ trav. & Scale \\
\midrule
LIVIN~\cite{LIVIN} & Real & Whole home & \cmark & \cmark & \cmark & \xmark & 30 homes \\
RoboCasa365~\cite{RoboCasa365} & Designed & Single room & \cmark & \cmark & \cmark & \xmark & 2.5K rooms \\
MolmoSpaces~\cite{MolmoSpaces} & Synthetic & Mixed & \cmark & \cmark & \pmark & \xmark & 230K+ homes / rooms \\
HomeWorld~\cite{HomeWorld} & Synthetic & Whole home & \pmark & \xmark & \xmark & \xmark & 5K homes \\
\textbf{RoboHome (ours)} & \textbf{RPLAN} & \textbf{Whole home} & \cmark & \cmark & \cmark & \cmark & \textbf{20K+ whole-home scenes} \\
\midrule
\multicolumn{8}{c}{\textbf{Part B: Recent scene generation methods}} \\
\midrule
Resource / Method & Floorplan source & Scope & Sim-ready & Interactive art. & Orig.\ scale & Emb.\ trav. & Layout approach \\
\midrule
SceneSmith~\cite{SceneSmith} & LLM-generated & Whole home & \cmark & \cmark & \xmark & \pmark & Multi-agent \\
SAGE~\cite{SAGE} & LLM-generated & Single room & \cmark & \pmark & \xmark & \xmark & Agent \\
EmbodiedGen V2~\cite{EmbodiedGen-V2} & Infinigen-generated & Whole home & \cmark & \xmark & \xmark & \xmark & Constraint solving + rescaling \\
HomeWorld~\cite{HomeWorld} & Fine-tuned LLM & Whole home & \pmark & \xmark & \xmark & \xmark & Image-generation placement \\
\textbf{RoboHome (ours)} & \textbf{RPLAN} & \textbf{Whole home} & \cmark & \cmark & \cmark & \cmark & \textbf{Envelope pre-layout + joint search} \\
\bottomrule
\end{tabular}}
\end{table*}

\begin{table*}[t]
\centering
\caption{Comparison of representative robot simulation datasets and data generation pipelines, ordered from tabletop manipulation to humanoid loco-manipulation and whole-home tasks. \cmark: supported; \xmark: not supported in the reported task suite; \pmark: partially supported; --: not reported.}
\label{tab:data_compare}
\setlength{\tabcolsep}{4pt}
\renewcommand{\arraystretch}{1.15}
\resizebox{0.96\textwidth}{!}{%
\begin{tabular}{@{}lllcccrr@{}}
\toprule
Dataset / Pipeline & Generation & Scene scope & Mobile & \makecell{Articulated\\manip.} & Multi-emb. & \#Tasks & \#Traj. \\
\midrule
LIBERO~\cite{LIBERO} & Teleoperation & Tabletop & \xmark & \cmark & \xmark & 130 & 6.5K \\
RoboTwin 2.0~\cite{RoboTwin2} & Rule-based generation & Tabletop & \xmark & \cmark & \cmark & 50 & 100K+ \\
\addlinespace[2pt]
HumanoidMimicGen~\cite{HumanoidMimicGen} & \makecell[l]{Demo. augmentation\\+ whole-body planning} & Factory workcell & \cmark & \xmark & \pmark$^{a}$ & 9 & 9K \\
SIMPLE~\cite{simple} & \makecell[l]{Motion planning\\/ VR teleoperation} & Indoor scenes & \cmark & \cmark & -- & 60 & 6K$^{b}$ \\
\midrule
\textbf{RoboHome (ours)} & \makecell[l]{\textbf{Video-guided}\\\textbf{automatic generation}} & \makecell[l]{\textbf{Real-floorplan}\\\textbf{whole home}} & \cmark & \cmark & \cmark & \textbf{6} & \textbf{60K+} \\
\bottomrule
\end{tabular}}
\par\smallskip{\footnotesize\raggedright $^{a}$G1 and a floating-base G1 variant. $^{b}$6K rendered episodes; SIMPLE provides both automatic and teleoperated data collection.\par}
\end{table*}

\section{Method}
\label{sec:method}

\begin{figure*}[t]
  \centering
  \includegraphics[width=0.82\textwidth]{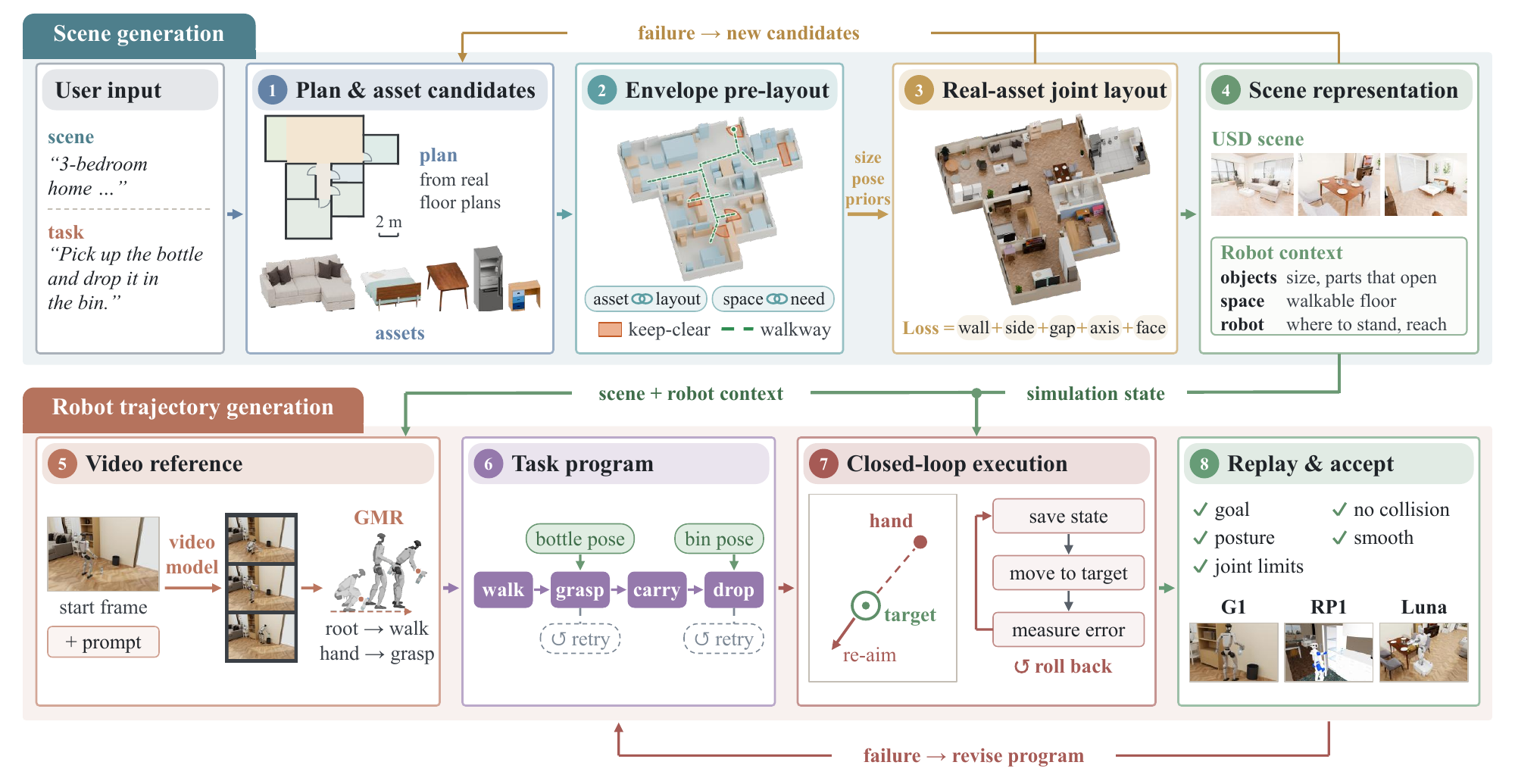}
  \caption{RoboHome pipeline: robot-ready homes from real floor plans (top) and robot trajectories from video (bottom).}
  \label{fig:pipeline}
\end{figure*}

\subsection{Problem Statement}
\label{sec:problem}

Beyond determining the home structure, the furniture assets, and their layout within limited space, scene generation for embodied long-horizon loco-manipulation must support cross-room robot traversal and continuous interaction. Drawing on residential design codes~\cite{GB50096,GB50352,GB55038} and traversal clearances, we follow four principles: (i) \emph{real structure and dimensions}: the home structure comes from normalized real floorplans and assets are never rescaled, so rooms are fitted only through asset selection and pose adjustment; (ii) \emph{function and task fit}: each room receives the basic furniture of its function, plus the objects and fixtures required by a given task; (iii) \emph{plausible layout and geometric validity}: assets are arranged as they are used, without out-of-bound placement or interpenetration and with valid support and mounting; and (iv) \emph{continuous traversal and usage space}: robot-traversable passages connect the entrance to every room, and usage space is reserved around furniture and task objects.

A scene description, a floorplan image, a task description, and a robot embodiment $B$ are optional inputs. We maintain a floorplan library $\mathcal{P}$ and an \emph{expert-annotated asset set} $\mathcal{A}$, qualified on eight annotated attributes including category, style, appearance, original size, functional front, and mounting conditions. Given a building shell $H$, an object list $F$, and a traversal configuration $\Gamma$, we seek a layout
\begin{equation}
E=\{(a_i,p_i)\}_{i\in\mathcal{I}},\ \ F_{\mathrm{req}}\subseteq\mathcal{I}\subseteq F,\ \ C(E;H,F,\Gamma)=1,
\label{eq:problem}
\end{equation}
where $\mathcal{I}$ is the set of placed objects, $a_i\in\mathcal{A}$ the selected asset, and $p_i=(x_i,y_i,z_i,\theta_i)$ its pose; the predicate $C$ requires geometric and mounting validity, the required relations, and the traversal and usage-space conditions. $H$ is built from the floorplan image, parsed by the open-source Qwen3-VL~\cite{Qwen3-VL}, or drawn from $\mathcal{P}$ when no image is provided. $F$ is built from the room functions and the scene description; a task requirement $Q=(O_q,M_q,I_q,Z_q)$ (task objects, their rooms, interaction type, and goal state) adds $O_q$ to $F$ within $M_q$ and determines the reserved operating regions. User-specified objects, task objects, and the essential furniture of each room form the required set $F_{\mathrm{req}}$, which the solver never omits. $\Gamma=(r_b,r_c,h_w,w_{\mathrm{main}},w_{\min})$ comprises the walking and carrying radii, the walking-height bound, and the main/minimum passage widths; these are assumed parameters, set from the embodiment when $B$ is given and otherwise to preset defaults (\eg, $r_b{=}0.175$\,m).

\subsection{Envelope Pre-Layout}
\label{sec:envelope}

Asset selection and layout are coupled, and furniture occupancy, operating space, and traversal compete for limited space, so locally feasible placements often fail the room-level traversal and validity checks. We therefore first lay out \emph{semantic envelopes} as proxies for the spatial requirements of objects, which mainly improves generation reliability (\cref{sec:ablation}), checking four types of constraints (\cref{fig:pipeline}).

\noindent\textbf{Constraints.}
\emph{Size constraints} ensure that an envelope can be instantiated by an asset: with $\mathcal{A}_i$ the candidate assets retrieved by category and appearance and $\mathbf{b}(a)$ the original size of asset $a$, a category prior size $\mathbf{d}\in\mathcal{D}^0_{c_i}$ is usable only if some $a\in\mathcal{A}_i$ satisfies $b_k(a)\le d_k+\epsilon$ for $k{=}1,2,3$ and $b_1(a)\,b_2(a)\ge \eta_{c_i} d_1 d_2$, where $\epsilon$ and $\eta_{c_i}$ are the size tolerance and the minimum fill ratio. \emph{Functional constraints} fix the pose of an object relative to its reference object (\eg, a nightstand beside its bed) through offsets determined by the relation type, both sizes, and a discretized distance; wall-mounted, corner, and supported objects must fit a wall, a corner, or a supporting surface. \emph{Spatial constraints} keep envelopes in bounds, non-overlapping, and out of reserved regions (door access areas, window fronts, usage zones of placed furniture, and the sweep, pull-out, or standing regions of task objects). \emph{Traversal constraints} project the occupancy within the height range $[0,h_w]$ onto the floor plane and dilate it by the traversal radius $r$, giving the free space of room $j$
\begin{equation}
\Omega_j^{\mathrm{free}}(r)=\Omega_j\setminus\big[(O_j\oplus K_r)\cup W_j(r)\big],
\label{eq:free}
\end{equation}
where $\Omega_j$ is the room region, $O_j$ the obstacle projection, $K_r$ the dilation kernel, and $W_j(r)$ the wall strip of width $r$ (excluding doorway fronts). All doorways of a room must be pairwise connected within this space, and a passage of width at least $w$ must lead from the entrance to the usage positions of beds, wardrobes, stoves, toilets, \etc. Doors are treated as open with their swing regions reserved.

\noindent\textbf{Solving.}
Each room is solved by depth-first search in dependency order over envelope sizes and poses. Traversal is checked after each floor-standing piece is placed; the search backtracks on failure and returns the first feasible solution within the budget, first with $(r_c,w_{\mathrm{main}})$ and, upon failure, with $(r_b,w_{\min})$. The envelopes pass size upper bounds and pose priors $\bar p_i$ to the instantiation stage: candidate assets are filtered to $\mathcal{A}^{\mathrm{env}}_i=\{a\in\mathcal{A}_i: b_k(a)\le d_{i,k}+\epsilon\}$, which is non-empty by the size constraint, so that the selected assets cannot encroach on space reserved for other objects and for traversal, while the priors only guide the search order.

\begin{figure*}[t]
  \centering
  \includegraphics[width=\textwidth]{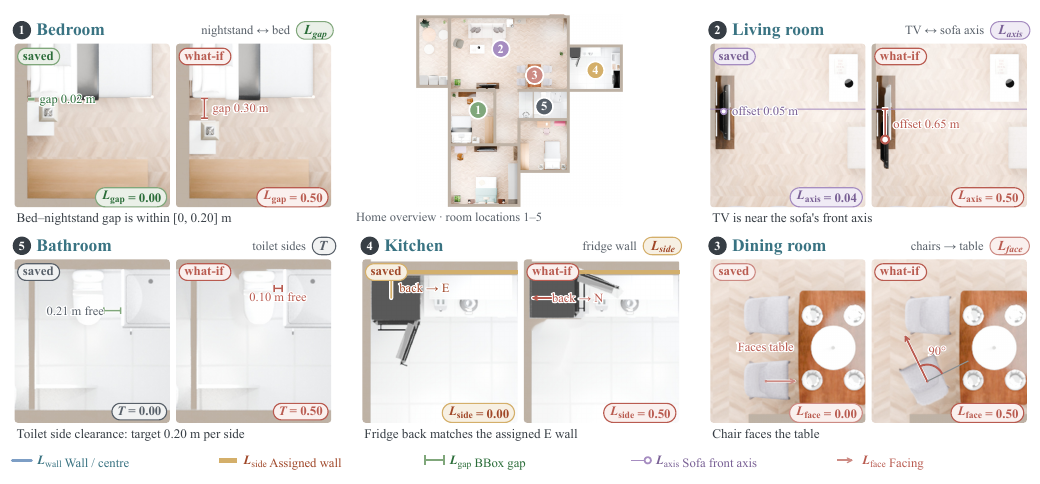}
  \caption{Room-specific layout rules. Saved layouts and illustrative pose perturbations (``what-if'') show bed--nightstand spacing ($L_{\mathrm{gap}}$), sofa--TV alignment ($L_{\mathrm{axis}}$), chair orientation ($L_{\mathrm{face}}$), assigned fridge wall ($L_{\mathrm{side}}$), and toilet side clearance (a functional-margin term). Lower penalties indicate better rule satisfaction. The legend also includes wall alignment/centering ($L_{\mathrm{wall}}$).}
  \label{fig:rules}
\end{figure*}

\subsection{Real-Asset Joint Layout}
\label{sec:joint}

The solver jointly searches assets and poses in $\mathcal{A}^{\mathrm{env}}_i$, ranking assets by size fit and group consistency and poses by
\begin{equation}
\begin{aligned}
\ell(p_i\mid a_i,Y_j^{<i}) ={}& L_{\mathrm{pri}}(p_i\mid \bar p_i)+\textstyle\sum_{k\in\mathcal{K}}\lambda_k L_k(p_i\mid a_i,Y_j^{<i})\\
&+\lambda_{\mathrm{mar}}L_{\mathrm{mar}}(Y_j^{<i}\cup\{p_i\}),
\end{aligned}
\label{eq:cost}
\end{equation}
where $Y_j^{<i}$ is the current partial layout, $L_{\mathrm{pri}}$ the deviation from the envelope prior, and $\mathcal{K}=\{\mathrm{wall},\allowbreak\mathrm{side},\allowbreak\mathrm{gap},\allowbreak\mathrm{axis},\allowbreak\mathrm{face}\}$ indexes five soft constraints (\cref{fig:rules}) that follow the constraint categories of GardenDesigner~\cite{GardenDesigner}, redefined after common residential layout rules: wall alignment or centering, the assigned wall side, furniture spacing, alignment of frontal axes, and functional orientation. Each term is enabled by furniture role and normalized, with fixed weights. $L_{\mathrm{mar}}(Y)=\sum_{m}\mu_m M_m(Y)$ is the functional-margin deficiency, with terms for clearances such as chair pull-out, bedside passages, and toilet side space; a length-type term is $M_m=\max(0,1-c/c^\star)$, the shortfall of the clearance $c$ relative to its target $c^\star$. We call the functional and spatial constraints and the cost $\ell$ the \emph{constraint metrics}.

Candidates must also satisfy the hard constraints, \ie, checks on boundaries, collisions, support and mounting, structural openings, required relations, and door-swing regions; on conflict, the search backtracks to adjust the pose or replace the asset. Since instance poses may deviate from the envelope priors, connectivity among the entrance door, the room doors, and the rooms is re-checked with $r_b$ after each floor-standing piece and after each room. After the first feasible room solution, the solver keeps searching within a limited budget and returns $Y_j^\star\in\arg\min_{Y\in\mathcal{V}_j}L_{\mathrm{mar}}(Y)$ over the visited feasible solutions $\mathcal{V}_j$, stopping early when $L_{\mathrm{mar}}=0$.

\subsection{Scene Representation and Output}
\label{sec:output}

After assembly, the system reads back the saved geometry and transforms and independently re-checks the object list, support, intersections, wall-mounting gaps, and structural openings; it also recomputes whole-home traversal with $r_b$ and reports blocking objects, and verifies architectural doors and jointed furniture for opening and joint motion in physics simulation. The final scene $S=(H,E,R,J,V)$ comprises the shell, the layout, the object relations $R$ and joint information $J$ derived from $E$ and the asset annotations, and the verification state $V$; it is output as a loadable USD scene with a per-object JSON manifest, so that task goals can be mapped to concrete entities. Given an embodiment $B$, the system further computes the interaction information $\mathcal{T}=\Phi(S,B)$: the traversable space of each room, the approach regions of objects, estimated end-effector reachability, and interference with part motion. These are geometric estimates; physical feasibility is verified only in trajectory generation (\cref{sec:trajgen}).

\section{Expert Trajectory Generation for Tasks}
\label{sec:trajgen}

Our trajectory pipeline requires no human demonstration (\cref{fig:pipeline}, bottom): a video generation model provides human-like motion references, and a program grounded in simulation ground truth ensures accurate end points and dynamic feasibility and is reused for batch production.

\noindent\textbf{Problem definition.}
Given the scene $S$ (\cref{sec:output}), the robot embodiment, and a task description $\ell$, the goal is to generate a long-horizon trajectory
\begin{equation}
\tau=\{a_t\}_{t=1}^{T},\qquad a_t=(p_t,R_t,q_t,c_t),
\label{eq:traj}
\end{equation}
where $a_t$ is the whole-body reference at 50\,Hz (for the Unitree G1, $a_t\in\mathbb{R}^{40}$: root position and orientation, 29 joint angles, and the opening of both hands), executed in physics simulation by a frozen whole-body tracking policy $\pi_{\mathrm{trk}}$~\cite{partymimic}. The trajectory must be (1) \emph{task-complete}: the final state satisfies the goal predicate set $\mathcal{G}$; (2) \emph{dynamically feasible}: it is executed by the tracking policy without interpenetration, dropped objects, or joint-limit violations; and (3) \emph{human-like}: the motion resembles how a person would perform it.

\noindent\textbf{1) Video reference.}
The agent reads $S$ and $\ell$ and writes the task goal $\mathcal{G}$ and a video prompt. The first frame containing the robot is rendered from a known camera pose $T_{\mathrm{cam}}$ and passed with the prompt to a video generation model~\cite{videomodel}, which produces a demonstration of the entire task. Whole-body motion is recovered through motion estimation and retargeting~\cite{motionest,GMR}, transformed into the scene frame with $T_{\mathrm{cam}}$, and segmented by root horizontal velocity into alternating locomotion and manipulation segments. Locomotion segments keep the planar root path as reference points; manipulation segments keep the gripper open/close timing and the hand path $\Delta_t$ relative to the segment end point, \ie, the motion shape with the landing point removed; the segment order and these quantities form the video reference.

\noindent\textbf{2) Code as Policies grounded in simulation ground truth.}
The agent writes the task as a Python program over an interface library $\mathcal{L}$~\cite{capx,codeaspolicies} of \emph{motion primitives} (walking, hand-path following, posture adjustment, hand opening and closing), \emph{ground-truth queries} (object poses, contacts, articulation states), and \emph{flow control} (state saving and restoring, conditional retries, per-frame failure checks). The primitives output the whole-body reference, whose sequence $\tau$ is the trajectory label. The program follows the segments of the video reference: locomotion segments call the walking primitive, whose kinematic motion generator passes through the reference points, and for manipulation segments the target point $x^*$ is read from simulation and the video hand path is translated as a whole so that its end point lands on $x^*$, \ie, $h_t=x^*+\Delta_t$, so that the shape of the path comes from the video and the landing point from simulation. Because the hand error of the tracking policy can reach several centimeters and exceed the grasping tolerance, the program saves the simulation state before each action and measures the actual hand position afterwards; if the deviation exceeds a threshold, it rolls back, compensates the end point by the opposite deviation, and re-executes. It also checks for toppling, dropped objects, and interpenetration at every frame and, on failure, retries from the start of the stage. The agent composes primitives into skills such as grasping and door opening and adds them to $\mathcal{L}$ for reuse in later tasks~\cite{voyager}.

\noindent\textbf{3) Batch generation.}
The expert program is rerun under randomized initial robot and object poses and trajectory perturbations, and the resulting reference is replayed open-loop from the same initial state to save the data; since the simulation is deterministic, the replay reproduces the executed result and serves as a repeatability check. A trajectory is kept if and only if its final state satisfies all predicates in $\mathcal{G}$ and it passes the checks on interpenetration, unexpected contact, posture, smoothness, and joint limits. Failed trajectories carry stage and cause codes, which the agent uses to revise the program. The video and the program are generated once per task, so their cost is amortized over all trajectories (\cref{sec:datagen_exp}).

\section{Dataset}
\label{sec:dataset}

With the two pipelines we build the \textbf{RoboHome} dataset (\cref{tab:scene_compare,tab:data_compare}). (1) The \textbf{whole-home interactive scene library} contains 20K+ original-scale scenes with context manifests, built from RPLAN floorplans~\cite{RPLAN} and \textbf{120K+} multi-source assets (HSSD~\cite{HSSD}, Objaverse~1.0~\cite{Objaverse}, ArtVIP~\cite{ArtVIP}, ReplicaCAD~\cite{Habitat2}, and YCB~\cite{YCB}). Of 80{,}788 source floorplans, 44{,}036 valid ones are retained (shell spot-check pass rate 596/600), and repair and pre-screening yield \textbf{23{,}221} usable models and \textbf{19{,}795} runtime-qualified assets. Scenes are exported to Isaac Sim~\cite{IsaacSim}; those with interpenetration, floating objects, or blocked doors are removed by post-save re-checks, and only articulated parts that pass the readback test are marked as interactive. (2) The \textbf{multi-embodiment long-horizon trajectory set} contains 60K+ trajectories (about 750 hours) of six loco-manipulation tasks for humanoid (Unitree G1, RP1) and wheeled dual-arm (Luna) robots, all validated by deterministic replay (\cref{sec:trajgen}). Following 3D scene graphs~\cite{3DSceneGraph}, scenes, instances~\cite{Habitat2}, and trajectories carry hierarchical structural, physical, and motion annotations. Training and test sets are separated by source floorplans and held-out seeds, and the dataset will be released after licensing review; construction and annotation details are given in the supplementary material.

\section{Experiments}
\label{sec:experiments}

\subsection{Experimental Setup}
\label{sec:setup}

\noindent\textbf{Test inputs and baselines.}
We use 20 inputs based on distinct RPLAN floorplans, each comprising a scene description (a text requirement, a structured object list, and room functions) and a floorplan image. We compare against Holodeck~\cite{Holodeck} and EmbodiedGen V2~\cite{EmbodiedGen-V2} under identical inputs; baselines use the modalities their interfaces accept (\eg, omitting the floorplan image when unsupported) and their native asset pools. RoboHome calls no commercial model for scene generation; its floorplan-image parsing uses the open-source Qwen3-VL~\cite{Qwen3-VL} run locally. Each method runs the 20 inputs with three initial seeds $s\in\{42,123,2026\}$ (input $k$ uses seed $s+k$, shared across methods), giving $N=20\times3{=}60$ attempts, each with a time budget $T_{\max}{=}90$\,min; timeouts count as failures.

\begin{figure*}[tb]
  \centering
  \includegraphics[width=0.80\textwidth]{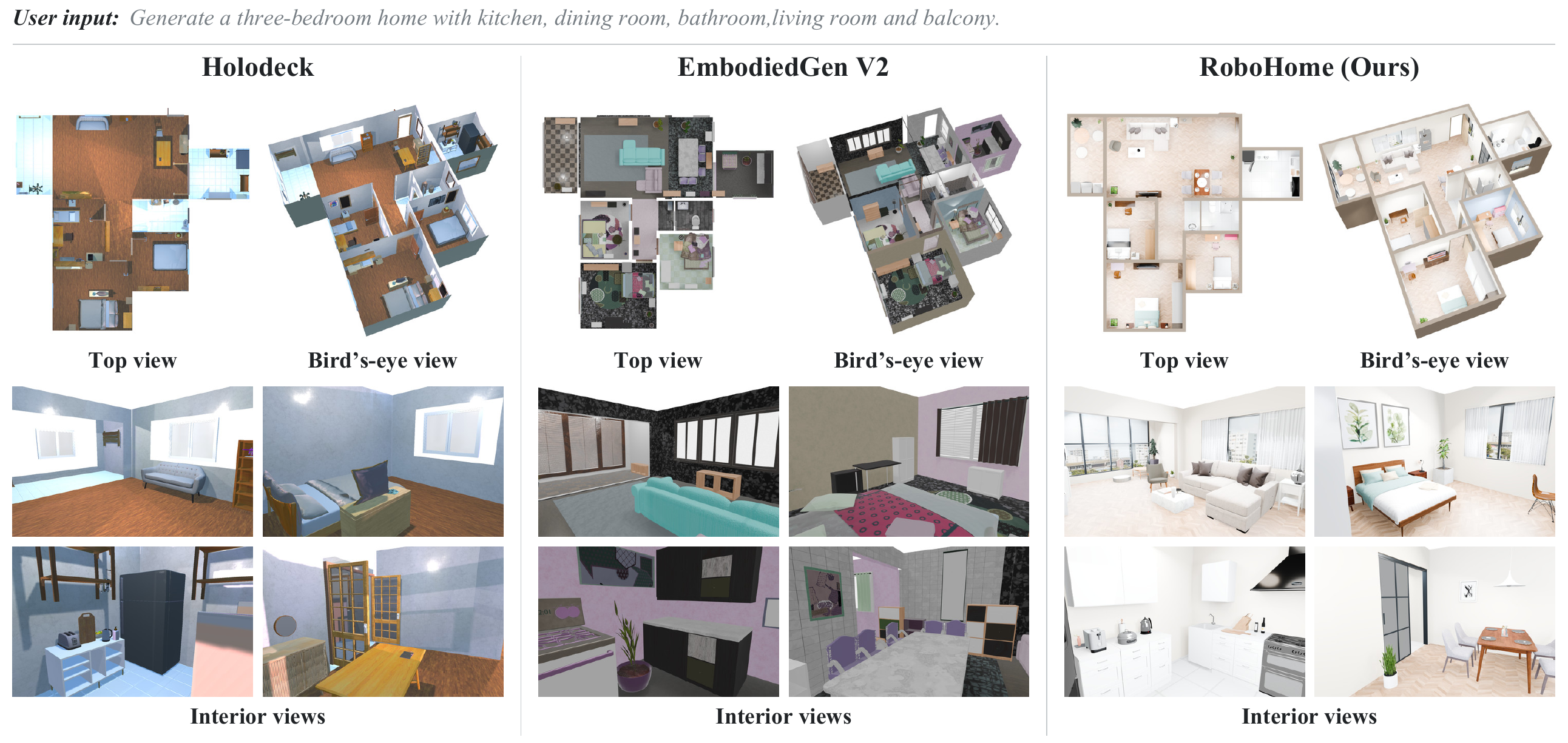}
  \caption{Qualitative comparison under the same input: top, bird's-eye, and four interior views per method (RoboHome: our scene generator).}
  \label{fig:qualitative}
\end{figure*}

\begin{table}[t]
\centering
\caption{\textbf{Objective metrics of whole-home scenes} (\%) over 60 attempts per method, averaged over generated scenes; SceneSmith~\cite{SceneSmith} generated none (0/60) and is omitted. A dimension score (italic) averages its metrics, lower-is-better ones complemented first. \textbf{Bold}: best; \underline{underline}: second best. $^*$Holodeck's native AI2-THOR assets partly lack collision/support evidence: $P_{\mathrm{col}}$ covers only the checkable subset, and objects without evidence count as invalid in $S_{\mathrm{sup}}$.}
\label{tab:objective}
\small
\setlength{\tabcolsep}{3pt}
\fitlinewidth{%
\begin{tabular}{@{}lccc@{}}
\toprule
Metric & Holodeck~\cite{Holodeck} & \makecell{EmbodiedGen\\V2~\cite{EmbodiedGen-V2}} & \textbf{Ours} \\
\midrule
Generated scenes (/60) & \underline{38} & 29 & \textbf{55} \\
\midrule
\emph{Functional completeness} & \underline{\emph{67.7}} & \emph{56.9} & \textbf{\emph{91.6}} \\
\quad Requirement coverage $C_{\mathrm{req}}\uparrow$ & \underline{76.3} & 58.9 & \textbf{91.7} \\
\quad Function coverage $C_{\mathrm{func}}\uparrow$ & \underline{59.1} & 54.9 & \textbf{91.6} \\
\midrule
\emph{Layout plausibility} & \underline{\emph{71.1}} & \emph{56.5} & \textbf{\emph{93.9}} \\
\quad Relation satisfaction $R_{\mathrm{rel}}\uparrow$ & \underline{71.1} & 56.5 & \textbf{93.9} \\
\midrule
\emph{Usability} & \underline{\emph{75.7}} & \emph{61.5} & \textbf{\emph{85.4}} \\
\quad Room reachability $R_{\mathrm{room}}\uparrow$ & \underline{95.5} & 70.9 & \textbf{97.7} \\
\quad Usage-zone reachability $R_{\mathrm{use}}\uparrow$ & \underline{55.9} & 52.2 & \textbf{73.2} \\
\midrule
\emph{Geometric validity} & \underline{\emph{85.6}} & \emph{84.9} & \textbf{\emph{99.1}} \\
\quad Collision rate $P_{\mathrm{col}}\downarrow$ & \underline{7.1}$^*$ & 15.9 & \textbf{0.2} \\
\quad Out-of-bound rate $P_{\mathrm{out}}\downarrow$ & \underline{0.8} & 6.3 & \textbf{0.0} \\
\quad Valid support $S_{\mathrm{sup}}\uparrow$ & 64.6$^*$ & \underline{76.9} & \textbf{97.6} \\
\bottomrule
\end{tabular}}

\end{table}

\begin{table}[t]
\centering
\caption{\textbf{Subjective visual evaluation} (1--5, $\uparrow$): three judge models (mean) and 59 volunteers; subscripts: std.}
\label{tab:subjective}
\small
\setlength{\tabcolsep}{4pt}
\fitlinewidth{%
\begin{tabular}{@{}llccc@{}}
\toprule
Dimension & Rater & Holodeck~\cite{Holodeck} & EmbodiedGen V2 & \textbf{RoboHome} \\
\midrule
\multirow{2}{*}{Visual aesthetics} & Models & \underline{2.43}$_{\pm0.22}$ & 2.14$_{\pm0.18}$ & \textbf{2.79}$_{\pm0.17}$ \\
 & Human & \underline{2.56}$_{\pm0.48}$ & 2.14$_{\pm0.35}$ & \textbf{2.65}$_{\pm0.48}$ \\
\midrule
\multirow{2}{*}{Visual realism} & Models & \underline{2.35}$_{\pm0.14}$ & 2.16$_{\pm0.14}$ & \textbf{2.60}$_{\pm0.23}$ \\
 & Human & \underline{2.26}$_{\pm0.45}$ & 2.03$_{\pm0.19}$ & \textbf{2.53}$_{\pm0.54}$ \\
\bottomrule
\end{tabular}}
\end{table}

\noindent\textbf{Evaluation metrics.}
Scene quality covers functional completeness ($C_{\mathrm{req}},C_{\mathrm{func}}$), layout plausibility ($R_{\mathrm{rel}}$), usability ($R_{\mathrm{room}},R_{\mathrm{use}}$), and geometric validity ($P_{\mathrm{col}},P_{\mathrm{out}},S_{\mathrm{sup}}$), using eight generator-independent metrics adapted from SceneEval~\cite{SceneEval} and PhyScene~\cite{PhyScene}. Requirements are frozen before generation; a match requires a distinct object with the correct room and asset-label category. GPT-6.1 Sol, GPT-6 Astra, and Claude Opus 5.5 each rate aesthetics and realism twice on a 1--5 scale from identical anonymized renderings. We report their mean, Kendall's $W$, and ICC, and cross-validate with Q-Align~\cite{QAlign} and 59 volunteers rating layout, aesthetics, and realism.

Yield measures attempts saving a simulator-loadable scene within $T_{\max}$; SR additionally requires all validation checks to pass, with missing evidence counted as failure. Checks cover preserved shell geometry and dimensions, collision-free in-bound placement, valid support/mounting, required items, unblocked doors, and entrance-to-room connectivity for radius $r{=}0.175$\,m. This radius is below the standing footprints of all three embodiments, so validation is necessary but insufficient for robot-specific feasibility. $T_{\mathrm{gen}}$ measures input-to-save time. Quality metrics cover generated scenes only; full definitions are in the supplement.

\subsection{Results and Analysis}
\label{sec:results}

\noindent\textbf{Scene quality.}
RoboHome succeeds in 55 of 60 attempts (SR 91.7\%) and outperforms both baselines in success rate and on all eight objective metrics (\cref{tab:objective}). Geometric validity shows the clearest advantage: its collision, out-of-bound, and valid-support rates are 0.2\%, 0\%, and 97.6\%, compared with 15.9\%, 6.3\%, and 76.9\% for EmbodiedGen V2. Functional completeness and layout plausibility exceed the runner-up Holodeck by 23.9 and 22.8 percentage points. Usage-zone reachability $R_{\mathrm{use}}$ remains our weakest metric at 73.2\% (55.9\% and 52.2\% for the baselines), against 97.7\% for room reachability: it also requires usage zones to fit a robot stance, so room-to-room passages alone are not sufficient.

\noindent\textbf{Qualitative comparison.}
Under the same input (\cref{fig:qualitative}), baseline layouts often combine local crowding with large empty areas, furniture orientations lack functional correspondence, overfilled rooms may omit required nightstands or dining chairs, and some objects interpenetrate or block doorways. RoboHome keeps furniture at its original dimensions along walls, groups sofas and TVs, dining tables and chairs, and beds and nightstands by function with consistent orientations, and connects room doors by passages, in agreement with $R_{\mathrm{rel}}$ (93.9\% \vs\ 71.1\% and 56.5\%).

\noindent\textbf{Visual and human perception evaluation.}
The three judge models agree on the method ranking (Kendall's $W{=}1.00$; scene-level ICC$(2,k)$ of 0.75 for aesthetics and 0.63 for realism) and rate RoboHome highest in both dimensions (\cref{tab:subjective}): 2.79 and 2.60 for aesthetics and realism, versus 2.43 and 2.35 for Holodeck and 2.14 and 2.16 for EmbodiedGen V2. The volunteers also rate RoboHome highest in layout plausibility and visual realism, exceeding Holodeck by 0.22 and 0.27 and EmbodiedGen V2 by 0.49 and 0.50, respectively; its aesthetic score (2.65) is comparable to Holodeck's (2.56).

\begin{figure*}[t]
  \centering
  \includegraphics[width=0.92\textwidth]{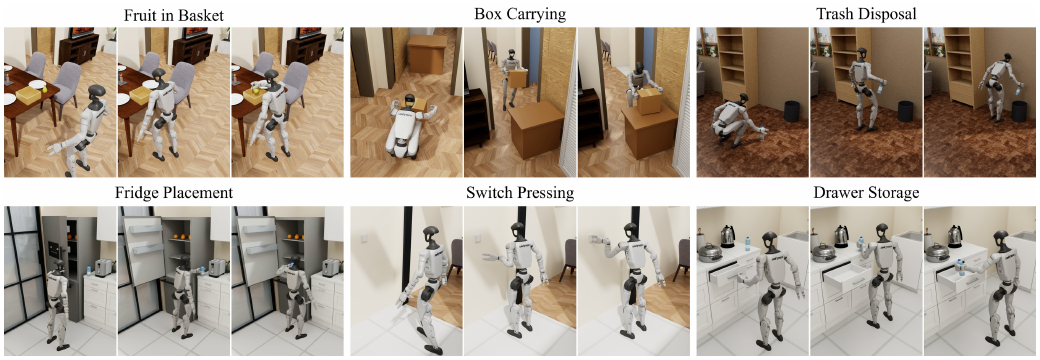}
  \caption{Unitree G1 performing six long-horizon loco-manipulation tasks in generated scenes (three keyframes per task).}
  \label{fig:tasks}
\end{figure*}

\subsection{Ablation Analysis}
\label{sec:ablation}

\begin{table}[t]
\centering
\caption{\textbf{Ablation study.} Yield and SR use 20 attempts; other metrics use scenes saved within the budget (\%; visual: 1--5). $T_{\mathrm{gen}}$: median time; $^\dagger$different host.}
\label{tab:ablation}
\scriptsize
\setlength{\tabcolsep}{2.2pt}
\fitlinewidth{%
\begin{tabular}{@{}lccccccccc@{}}
\toprule
Config. & Func. & Layout & Usab. & Geom. & Aes. & Real. & Yield$\uparrow$ & SR$\uparrow$ & $T_{\mathrm{gen}}$\,(s)$\downarrow$ \\
\midrule
Full method & \textbf{92.0} & \textbf{95.4} & \textbf{87.8} & \textbf{99.4} & 2.77 & \underline{2.61} & \textbf{95.0} & \textbf{95.0} & 839 \\
w/o envelope pre-layout & \underline{91.6} & \underline{94.3} & \underline{86.9} & \underline{99.0} & \textbf{2.79} & \textbf{2.63} & \underline{85.0} & 40.0 & \underline{723} \\
w/o constraint metrics & \textbf{92.0} & 42.8 & 26.6 & \underline{99.0} & 2.73 & 2.60 & \textbf{95.0} & 25.0 & \textbf{260} \\
w/o expert-annot.\ assets & 85.4 & 87.5 & 78.5 & 97.6 & \underline{2.78} & 2.57 & 75.0 & \underline{70.0} & 2149$^\dagger$ \\
\bottomrule
\end{tabular}}
\end{table}

We evaluate three variants on the same 20 inputs with initial seed $s{=}42$; inputs, candidate assets, evaluators, and time budgets are fixed except for the ablated component (\cref{tab:ablation}; qualitative results: supplementary material).

\noindent\textbf{Envelope pre-layout.}
Without pre-layout, the joint asset--pose search runs directly. Median generation time falls from 839\,s to 723\,s and scene quality stays similar, but yield falls from 95\% to 85\% and SR from 95\% to 40\%: pre-layout mainly improves generation reliability, not layout quality.

\noindent\textbf{Constraint metrics.}
Without \emph{constraint metrics}, the spatial and functional constraints (\cref{sec:envelope}) and the cost $\ell$ (\cref{eq:cost}) are removed, keeping only boundary and collision checks. Yield stays at 95\%, yet layout plausibility drops from 95.4\% to 42.8\%, usability from 87.8\% to 26.6\%, and SR to 25\%; relations become implausible, \eg, a TV blocking a doorway. Median time decreases to 260\,s, so constraint search accounts for much of the runtime but is central to scene usability. Geometric validity stays at 99.0\% (99.4\% for the full method), so it alone does not capture the spatial organization that robots need.

\noindent\textbf{Expert-annotated asset set.}
Bypassing expert qualification reduces yield to 75\%, SR to 70\%, layout plausibility to 87.5\%, and usability to 78.5\%. Its runtime (2149\,s) is not comparable because it used a different host.

\begin{table}[t]
\centering
\caption{Data generation and policy success (20 trials) on six G1 tasks. $Y$: held-out pass rate; $C$: amortized expert-program API cost; $L$: time per passing trajectory.}
\label{tab:datagen}
\small
\setlength{\tabcolsep}{3.5pt}
\fitlinewidth{%
\begin{tabular}{@{}lcccccc@{}}
\toprule
& \multicolumn{3}{c}{Generation} & \multicolumn{3}{c}{Policy success (\%)} \\
\cmidrule(lr){2-4}\cmidrule(lr){5-7}
Task & $Y$ (\%) & $C$ (\$) & $L$ (s) & ACT & $\pi_{0.5}$ & GR00T N1.7 \\
\midrule
Fruit in basket & 100 & 0.2 & 72 & 15 & 55 & 60 \\
Box carrying & 86 & 1.6 & 154 & 5 & 20 & 35 \\
Trash disposal & 100 & 0.5 & 75 & 10 & 45 & 75 \\
Fridge placement & 67 & 1.6 & 246 & 0 & 15 & 35 \\
Switch pressing & 96 & 0.3 & 50 & 20 & 60 & 75 \\
Drawer storage & 75 & 1.3 & 205 & 0 & 20 & 40 \\
\bottomrule
\end{tabular}}
\end{table}

\subsection{Data Generation and Policy Evaluation}
\label{sec:datagen_exp}

\noindent\textbf{Setup.}
We generate six long-horizon loco-manipulation tasks (\cref{fig:tasks,tab:datagen}), covering grasping, carrying, pressing, doors, and drawers. The dataset spans G1, RP1, and Luna (60K+ trajectories, \cref{tab:data_compare}); quantitative evaluation uses G1.

\noindent\textbf{Generation efficiency.}
\Cref{tab:datagen} reports the pass rate $Y$ on held-out seeds, the per-trajectory cost $C$ (the coding agent's token usage for writing the expert program at public Claude Opus 5.5 prices, amortized over the 50 accepted trajectories per task), and the single-process generation time $L$ per passing trajectory, including failed attempts. Across tasks, $Y$ averages 87.3\% (67\%--100\%), $C$ \$0.92, and $L$ 134\,s. API cost is incurred only while synthesizing and refining each expert program; finalized programs then generate further trajectories in parallel without API calls or human intervention. Amortized over the 300 fine-tuning trajectories (approximately 3.75 hours of data), the program-authoring API cost is about \$70 per hour of trajectory data; this covers API expenditure only, not video generation, simulation, or scene generation.

\noindent\textbf{Policy evaluation.}
We fine-tune ACT~\cite{ACT}, $\pi_{0.5}$~\cite{pi05}, and GR00T N1.7~\cite{gr00t} on 50 passing trajectories per task; all output the same 40-dimensional whole-body reference action, executed by a frozen tracking policy. Closed-loop evaluation uses 20 trials per task with unseen seeds and the same judge as trajectory generation. GR00T N1.7 has the highest observed success rate on every task, followed by $\pi_{0.5}$ and ACT (\cref{tab:datagen}); ACT records no success on fridge placement or drawer storage, while $\pi_{0.5}$ and GR00T N1.7 succeed on all six tasks. A high pass rate does not imply easy learning: box carrying has an 86\% pass rate but at most 35\% policy success. With only 50 demonstrations per task, these results give preliminary evidence that the generated data support loco-manipulation policy training in simulation, while loco-manipulation and articulated interaction remain challenging.

\section{Conclusion and Limitations}
\label{sec:conclusion}
RoboHome combines semantic-envelope pre-layout and joint asset--pose search to generate whole-home scenes, and coding-agent-written programs to synthesize loco-manipulation trajectories. Experiments demonstrate improved scene quality and reliability and support policy fine-tuning in simulation.

Limitations include asset coverage, rendering quality, single-level RPLAN homes, and geometric interaction estimates requiring trajectory verification. Quantitative trajectory and policy evaluation covers only Unitree G1 in one simulator. Future work will expand assets, embodiments, tasks, and indoor domains.

\section*{Statements}
\phantomsection
\addcontentsline{toc}{section}{Statements}
\statement{LLM usage}{Qwen3-VL~\cite{Qwen3-VL} parses floorplans locally; Seedance 2.0~\cite{videomodel} generates reference videos; Claude Opus 5.5 writes expert programs. GPT-6.1 Sol, GPT-6 Astra, and Claude Opus 5.5 judge renderings; baselines use GPT-5.6 Sol. Codex and Claude assisted literature organization, editing, and engineering. The authors reviewed all content and take responsibility for the paper.}

\renewcommand{\refname}{References}
\phantomsection
\addcontentsline{toc}{section}{References}

\bibliographystyle{plainnat}
\bibliography{robohome-refs,supplementary-refs}
\clearpage

\section*{Author List}
\phantomsection
\addcontentsline{toc}{section}{Author List}
\begingroup
\setlength{\columnsep}{12mm}
\begin{multicols}{2}
\raggedcolumns
\textbf{Project Leader}
\begin{itemize}[leftmargin=0pt]
    \item[] Haifeng Zhang
\end{itemize}

\medskip
\textbf{Core Contributors}
\begin{itemize}[leftmargin=0pt]
    \item[] Yuchen Wang
    \item[] Yuchen Bian
    \item[] Ningze Qi
    \item[] Shengze Xu
    \item[] Qizhao Geng
\end{itemize}

\medskip
\textbf{Contributors}
\begin{itemize}[leftmargin=0pt]
    \item[] Haoyang Luo
    \item[] Hanbing Zhang
    \item[] Qimai Yan
    \item[] Zhexiong Wang
    \item[] Zengxiang Huang
    \item[] Rui Song
    \item[] Jianjie Chen
    \item[] Guangxin Yang
    \item[] Junhan Sun
\end{itemize}

\columnbreak
\textbf{Corresponding Author}
\begin{itemize}[leftmargin=0pt]
    \item[] Qiaozhi Wang\\
        \href{mailto:wangqiaozhi@roboparty.com}{\nolinkurl{wangqiaozhi@roboparty.com}}
\end{itemize}
\end{multicols}
\endgroup

\clearpage

\reportcontents
\clearpage

\appendix
\section*{Supplementary Material}
\phantomsection
\addcontentsline{toc}{section}{Supplementary Material}
\setcounter{figure}{0}
\setcounter{table}{0}
\setcounter{equation}{0}
\renewcommand{\thefigure}{S\arabic{figure}}
\renewcommand{\thetable}{S\arabic{table}}
\renewcommand{\theequation}{S\arabic{equation}}
\renewcommand{\theHfigure}{S\arabic{figure}}
\renewcommand{\theHtable}{S\arabic{table}}
\renewcommand{\theHequation}{S\arabic{equation}}

This supplement provides the details that the main paper defers for space.
\Cref{sec:s_scene} expands RoboHome, our scene generator (Sec.~3 of the main paper), with the full constraint set, the soft-constraint and functional-margin terms, and the solver configuration.
\Cref{sec:s_traj} details the expert trajectory generation pipeline (Sec.~4).
\Cref{sec:s_data} gives the dataset construction, annotation, and validation details (Sec.~5).
\Cref{sec:s_metrics} gives the full evaluation protocol, metric definitions and thresholds, and the relation between the acceptance checks behind SR and the scores of Tab.~3 (Sec.~6.1).
\Cref{sec:s_judges} lists the per-judge visual scores behind Tab.~4 and the Q-Align results.
\Cref{sec:s_ablation} gives the ablation configurations and qualitative ablation results (Sec.~6.3).
\Cref{sec:s_tasks} specifies the six long-horizon tasks and the accounting behind the data generation results (Sec.~6.4).
Section, table, figure, and equation numbers without the prefix ``S'' refer to the main paper. We use \emph{RoboHome} consistently for the scene generation method, the framework, the released scene library, and the trajectory dataset.

\section{Scene Generation Details}
\label{sec:s_scene}
This section complements Sec.~3 of the main paper and details RoboHome, our scene generation method. We restate the notation only where needed and give the parts of the formulation that the main paper summarizes in prose. Every stage serves the same goal: a generated home must be directly usable for robot loco-manipulation, \ie, a robot can enter at the entrance, walk to every room, stand where each object is used, and open doors and drawers without leaving the original proportions of a real home.

\subsection{Inputs, Building Shell, and Object List}
\label{sec:s_inputs}
A scene description, a floorplan image, a task description, and a robot embodiment $B$ are optional inputs (Sec.~3.1).
The building shell $H$ is built from a metric floorplan $P$. When a floorplan image is given, it is parsed by the open-source Qwen3-VL~\cite{Qwen3-VL}, run locally, and normalized to $P$; otherwise $P$ is drawn from the floorplan library $\mathcal{P}$ according to the scene description, or at random when the description does not constrain it. The library $\mathcal{P}$ consists of normalized RPLAN floorplans~\cite{RPLAN} (\cref{sec:s_data}).

The object list $F$ is built by rules from the room functions and the scene description; no commercial model is called in scene generation. A task requirement $Q=(O_q,M_q,I_q,Z_q)$ lists the task objects, their rooms, the interaction type, and the goal state: $O_q$ is added to $F$ and restricted to the rooms $M_q$, and $(I_q,Z_q)$ determine the operating regions reserved for the task. User-specified objects, task objects, and the essential furniture of each room form the required set $F_{\mathrm{req}}$, which the solver never omits; the required relations come from functional rules and from the task.

The traversal configuration $\Gamma=(r_b,r_c,h_w,w_{\mathrm{main}},w_{\min})$ comprises the walking radius, the carrying radius, the upper bound of the walking height, and the main and minimum passage widths. It also fixes the open/closed state assumed for doors. $\Gamma$ is set from the embodiment dimensions when $B$ is given and otherwise takes preset defaults (\eg, $r_b{=}0.175$\,m), which yields general-purpose scenes. Because $\Gamma$ enters every traversal check below, the same floorplan can be furnished for a slim humanoid or for a wider wheeled base.

\subsection{Envelope Pre-Layout}
\label{sec:s_envelope}
Pre-layout addresses two couplings. Asset dimensions limit the feasible positions while the limited space restricts asset selection; and furniture occupancy, operating space, and traversal compete for the same floor area, so locally feasible placements rarely guarantee whole-room connectivity and task interaction space. Semantic envelopes therefore act as proxies for the spatial requirements of objects, and four types of constraints are checked on them.

\noindent\textbf{Size constraints.}
Let $\mathcal{A}_i$ be the candidate assets retrieved for object $i$ by category and appearance, and $\mathbf{b}(a)$ the original size of asset $a$. Because assets are never rescaled, a category prior size $\mathbf{d}\in\mathcal{D}^0_{c_i}$ is usable only if some $a\in\mathcal{A}_i$ satisfies
\begin{equation}
b_k(a)\le d_k+\epsilon\ \ (k=1,2,3),\qquad b_1(a)\,b_2(a)\ge \eta_{c_i}\, d_1 d_2,
\label{eq:s_size}
\end{equation}
where $\epsilon$ is the size tolerance and $\eta_{c_i}$ the minimum fill ratio of category $c_i$. If an asset is larger than all prior sizes of its category, its own size is added as an extra candidate size, so that large but valid assets are not excluded by the priors. Keeping assets at original scale keeps countertop heights, corridor widths, and the space in front of cabinets consistent with real homes, so the reach and clearance a robot experiences in simulation follow real proportions.

\noindent\textbf{Functional constraints.}
An object $i$ with a reference object $\rho(i)$ (\eg, a nightstand and its bed) must satisfy
\begin{equation}
\mathbf{c}_i=\mathbf{c}_{\rho(i)}+\alpha_i\mathbf{f}_{\rho(i)}+\beta_i\mathbf{l}_{\rho(i)},\qquad
\theta_i=\theta_{\rho(i)}+\Delta\theta_i,
\label{eq:s_function}
\end{equation}
where $\mathbf{c}$ is the planar center, $\theta$ the yaw, and $\mathbf{f},\mathbf{l}$ the forward and lateral unit vectors of the reference object. The offsets $\alpha_i,\beta_i,\Delta\theta_i$ are determined by the relation type, the sizes of both objects, and a discretized distance. Wall-mounted, corner, and supported objects must fit against a wall, into a corner, or on a supporting surface, respectively.

\noindent\textbf{Spatial constraints.}
Envelopes must stay within the room, must not overlap, and must not intrude into the reserved regions $\mathcal{R}_j$ of room $j$. $\mathcal{R}_j$ contains the access area on the inner side of each door, the area in front of windows, the front usage zones of already placed furniture (side access zones for beds), and the sweep, pull-out, or standing regions reserved for task objects according to $(I_q,Z_q)$. These regions are where a robot stands or where a door or drawer moves while it is operated, so keeping them free at the envelope stage prevents layouts in which an object is present but cannot be used.

\noindent\textbf{Traversal constraints.}
The occupancy within the height range $[0,h_w]$ is projected onto the floor and dilated by the traversal radius $r$, which gives the free space $\Omega_j^{\mathrm{free}}(r)$ of Eq.~(2). All doorways of a multi-door room must be pairwise connected within $\Omega_j^{\mathrm{free}}(r)$, and a passage of width at least $w$ must lead from the entrance to the usage positions of beds, wardrobes, stoves, toilets, and similar fixtures. Doors are treated as open with their swing regions reserved; the stances needed to open doors are handled in trajectory generation (Sec.~4).

\noindent\textbf{Solving.}
Each room is solved by depth-first search in dependency order over envelope sizes and poses. Pose candidates are tried in the priority order of orientation relations, wall alignment, and corner alignment. A candidate must satisfy all four constraint types; the traversal constraint is checked after each floor-standing piece is placed. If the candidate passes, the reserved region of its envelope is recorded; otherwise the search backtracks. The search returns the first feasible solution within the budget, first with $(r_c,w_{\mathrm{main}})$ and, upon failure, with $(r_b,w_{\min})$.

\noindent\textbf{Interface to the joint layout.}
Each envelope passes a size upper bound $\mathbf{d}_i$ and a pose prior $\bar p_i$ (center, orientation, and wall side) to the instantiation stage. The candidate assets are filtered to
\begin{equation}
\mathcal{A}^{\mathrm{env}}_i=\big\{a\in\mathcal{A}_i:\ b_k(a)\le d_{i,k}+\epsilon,\ k=1,2,3\big\},
\label{eq:s_filter}
\end{equation}
which is non-empty by \cref{eq:s_size}. The upper bounds keep the selected assets from encroaching on space reserved for other objects and for traversal, whereas the pose priors only order the search: the actual asset geometry is authoritative. Embedded objects pass only their centers and orientations. If the envelope stage of a room has no solution, no prior is passed and the joint search runs directly on the full candidate set $\mathcal{A}_i$. The ``w/o envelope pre-layout'' ablation (\cref{sec:s_ablation}) removes this stage for all rooms.

\subsection{Real-Asset Joint Layout}
\label{sec:s_joint}
The solver jointly searches assets in $\mathcal{A}^{\mathrm{env}}_i$ and their poses. Assets are ranked by size fit and group consistency: tables and chairs, as well as bed groups, prefer assets from the same source. Pose candidates are ranked by the cost $\ell$ of Eq.~(3), whose soft terms are summarized in \cref{tab:s_soft}. Each term is enabled according to the role of the furniture piece, its violation is normalized, and the weights $\lambda_k$ and $\lambda_{\mathrm{mar}}$ are set manually and then kept fixed.

\begin{table}[t]
\centering
\caption{Soft constraints in the pose cost $\ell$ (Eq.~(3)). Examples are those of Fig.~3; all terms are normalized and lower is better.}
\label{tab:s_soft}
\small
\setlength{\tabcolsep}{3pt}
\begin{tabular}{@{}l>{\raggedright\arraybackslash}p{0.33\columnwidth}>{\raggedright\arraybackslash}p{0.47\columnwidth}@{}}
\toprule
Term & Rule & Example (Fig.~3) \\
\midrule
$L_{\mathrm{wall}}$ & Wall alignment or centering & -- (legend of Fig.~3) \\
$L_{\mathrm{side}}$ & Assigned wall side & Fridge back matches the assigned wall \\
$L_{\mathrm{gap}}$ & Spacing between furniture & Bed--nightstand gap within $[0,0.20]$\,m \\
$L_{\mathrm{axis}}$ & Alignment of frontal axes & TV near the sofa's front axis \\
$L_{\mathrm{face}}$ & Functional orientation & Chair faces the table \\
\midrule
$L_{\mathrm{mar}}$ & Functional-margin deficiency & Toilet side clearance, target 0.20\,m per side \\
\bottomrule
\end{tabular}
\end{table}

\noindent\textbf{Functional-margin deficiency.}
The term $L_{\mathrm{mar}}(Y)=\sum_{m}\mu_m M_m(Y)$ is evaluated on the (partial) room layout $Y$. Its terms cover seating combinations, chair pull-out space, the space in front of armchairs, toilet side clearance, bedside passages, and the serving section. A length-type term measures the shortfall of a clearance $c$ relative to its target $c^\star$,
\begin{equation}
M_m=\psi(c;c^\star)=\max\big(0,\,1-c/c^\star\big),
\label{eq:s_margin}
\end{equation}
so it is zero once the target is met and grows linearly as the clearance shrinks. These clearances are the approach spaces a person or a robot needs in order to sit, reach, or pass beside furniture.

\noindent\textbf{Hard constraints and connectivity.}
Every candidate must pass the hard checks on boundaries, collisions, support and mounting, structural openings, required relations, and door-swing regions; on conflict, the search backtracks to adjust the pose or to replace the asset. Because instance poses may deviate from the envelope priors, connectivity among the entrance door, the room doors, and the rooms is re-checked with $r_b$ after each floor-standing piece and again for all objects after each room is completed.

\noindent\textbf{Solution selection.}
After the first feasible solution of room $j$ is found, the solver continues within a limited budget and returns $Y_j^\star\in\arg\min_{Y\in\mathcal{V}_j}L_{\mathrm{mar}}(Y)$, where $\mathcal{V}_j$ is the set of visited complete feasible solutions including the first one; the search stops early when $L_{\mathrm{mar}}{=}0$. $L_{\mathrm{mar}}$ thus both ranks candidate poses (through $\ell$) and selects among feasible room solutions.

\noindent\textbf{Search configuration.}
The envelope pre-layout uses 5\,cm voxels. The joint asset--pose search uses a 0.4\,m pose grid and expands at most 80{,}000 search nodes per room. The grid is only one source of pose candidates: candidates are also generated from rule relations (\cref{eq:s_function}) and from key geometric positions such as walls and corners, so clearances finer than the grid spacing can still be met.

\subsection{Scene Output and Interaction Information}
\label{sec:s_output}
After assembly, the system reads back the saved geometry and transforms and independently re-checks the object list, support, intersections, wall-mounting gaps, and structural openings. It recomputes whole-home traversal with $r_b$ and reports blocking objects, and it verifies architectural doors and furniture with original joints for opening and joint motion in physics simulation. The final scene $S=(H,E,R,J,V)$ comprises the shell, the layout, the object relations $R$ and joint information $J$ derived from $E$ and the asset annotations, and the verification state $V$. It is exported as a loadable USD scene with a per-object JSON manifest, so that task goals can be mapped to concrete entities.

Given an embodiment $B$, the interaction information $\mathcal{T}=\Phi(S,B)$ contains (i) the traversable space of each room under the embodiment's $\Gamma$; (ii) the approach regions of objects, \ie, candidate standing regions for using each object; (iii) estimated end-effector reachability from these regions; and (iv) interference between the robot and the motion of articulated parts, such as door sweeps (\cref{fig:s_traj}, top). All four are geometric estimates computed from $S$ and the dimensions of $B$. They are used to set up trajectory generation (\cref{sec:s_traj}), but they are not evidence of feasibility: whether the robot can actually reach, grasp, and operate an object is established only by physics execution and replay in Sec.~4.

\section{Expert Trajectory Generation Details}
\label{sec:s_traj}
\begin{figure*}[t]
  \centering
  \includegraphics[width=0.95\textwidth]{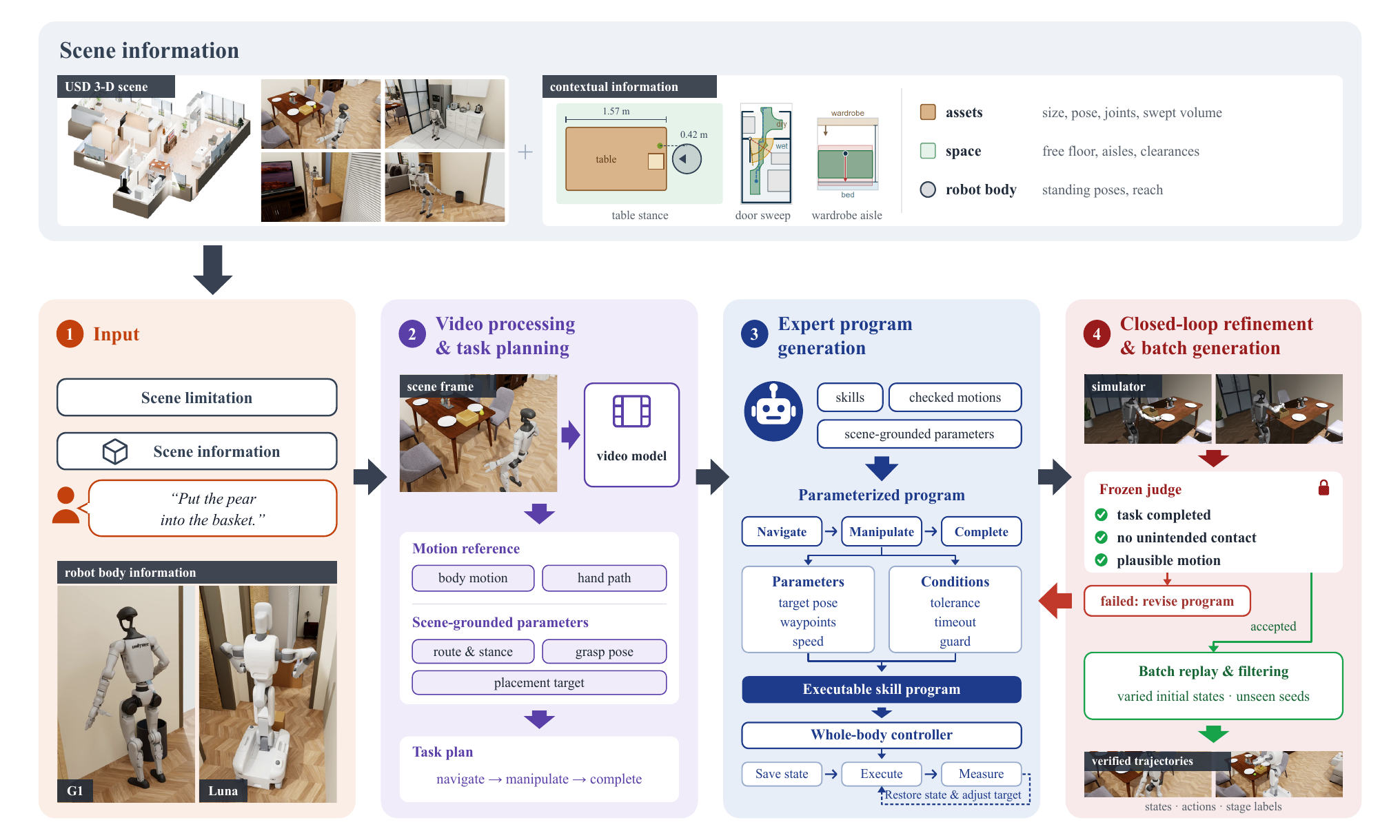}
  \caption{\textbf{Detailed view of expert trajectory generation.}
  \emph{Top:} the scene information passed to trajectory generation, \ie, the USD scene with its renderings and contextual information on assets (size, pose, joints, swept volume), space (free floor, aisles, clearances), and the robot body (standing poses, reach); the insets illustrate a table stance, a door sweep, and a wardrobe aisle.
  \emph{(1) Input:} scene information, scene limitations, a task description (here, ``Put the pear into the basket.''), and the robot body (G1 and Luna shown).
  \emph{(2) Video processing and task planning:} the scene frame is passed to the video model to obtain the motion reference (body motion and hand path); scene-grounded parameters (route and stance, grasp pose, placement target) and the task plan (navigate $\rightarrow$ manipulate $\rightarrow$ complete) are derived alongside.
  \emph{(3) Expert program generation:} the coding agent composes skills and checked motions into a parameterized program with per-stage parameters (target pose, waypoints, speed) and conditions (tolerance, timeout, guard); the whole-body controller executes it with save-state, execute, measure, and restore-and-adjust steps.
  \emph{(4) Closed-loop refinement and batch generation:} a frozen judge accepts a rollout only if the task is completed without unintended contact and with plausible motion; failures return to program revision, and accepted programs are replayed in batch under varied initial states and unseen seeds to produce verified trajectories with states, actions, and stage labels.}
  \label{fig:s_traj}
\end{figure*}

This section complements Sec.~4 of the main paper. \Cref{fig:s_traj} expands the bottom half of Fig.~2 and shows how the scene information of \cref{sec:s_output} enters trajectory generation.

\noindent\textbf{Trajectory and execution.}
A trajectory is the sequence $\tau=\{a_t\}_{t=1}^{T}$ of whole-body references at 50\,Hz (Eq.~(4)). For the Unitree G1, $a_t\in\mathbb{R}^{40}$ contains the root position and orientation, 29 joint angles, and the opening of both hands. The reference is executed in physics simulation by the frozen whole-body tracking policy $\pi_{\mathrm{trk}}$~\cite{partymimic}; the action label stored with each trajectory is this commanded reference. The same pipeline is applied to the RoboParty RP1 and the wheeled dual-arm Luna~\cite{GentoLuna} with their own embodiment models; the quantitative results in the main paper use G1 only.

\noindent\textbf{Video reference.}
The coding agent reads the scene $S$ and the task description and writes the goal predicate set $G$ and a video prompt. The first frame, which contains the robot, is rendered from a known camera pose $T_{\mathrm{cam}}$ and passed with the prompt to the video generation model (Seedance~2.0~\cite{videomodel}), which produces a demonstration of the entire task. Whole-body motion is recovered by motion estimation and retargeting~\cite{GMR,motionest} and transformed into the scene frame with $T_{\mathrm{cam}}$. The motion is split by root horizontal velocity into alternating locomotion and manipulation segments. A locomotion segment keeps only its planar root path, sampled into reference points. A manipulation segment keeps the gripper open/close timing and the hand path relative to the segment end point, $\Delta_t=h^{\mathrm{vid}}_t-h^{\mathrm{vid}}_{\mathrm{end}}$, \ie, the shape of the motion without its landing point. The ordered segments and these quantities form the video reference.

\noindent\textbf{Interface library.}
The expert program is a Python program over an interface library $\mathcal{L}$~\cite{codeaspolicies,capx} with three groups of functions (\cref{tab:s_interface}). Every motion primitive outputs the whole-body reference, so the sequence of primitive outputs is the trajectory label $\tau$. The agent composes primitives into skills such as grasping and door opening and adds them to $\mathcal{L}$ for reuse in later tasks~\cite{voyager}.

\begin{table}[t]
\centering
\caption{Interface library $\mathcal{L}$ used by expert programs.}
\label{tab:s_interface}
\small
\begin{tabular}{@{}lp{0.62\columnwidth}@{}}
\toprule
Group & Functions \\
\midrule
Motion primitives & Walking; hand-path following; posture adjustment; hand opening and closing \\
Ground-truth queries & Object poses; contact relations; articulation states \\
Flow control & Saving and restoring the simulation state; conditional retries; per-frame failure checks \\
\midrule
Composed skills & \eg, grasping, door opening (added to $\mathcal{L}$ by the agent) \\
\bottomrule
\end{tabular}
\end{table}

\noindent\textbf{Grounding motion in simulation.}
The program follows the segments of the video reference. A locomotion segment calls the walking primitive, whose kinematic motion generator produces a whole-body reference through the reference points. For a manipulation segment, the target point $x^\ast$ (\eg, a grasp point or a handle) is read from the simulation ground truth, and the video hand path is translated so that it ends at $x^\ast$:
\begin{equation}
h_t=x^\ast+\Delta_t .
\label{eq:s_anchor}
\end{equation}
The shape of the path thus comes from the video and its landing point from the simulation.

\noindent\textbf{Rollback and compensation.}
The hand error of the tracking policy can reach several centimeters and exceed the grasping tolerance. Before each action, the program saves the simulation state; after the action, it measures the actual hand position $\hat h$ and the deviation $e=\hat h-x^\ast$. If $\|e\|$ exceeds a threshold, the program restores the saved state, shifts the commanded end point to $x^\ast-e$, and re-executes the action. Independently, it checks for toppling, dropped objects, and interpenetration at every frame and, on failure, restores the state saved at the start of the stage and retries the stage.

\noindent\textbf{Batch generation and acceptance.}
For batch production, the finalized expert program is rerun under randomized initial robot and object poses and trajectory perturbations. The resulting reference is then replayed open-loop from the same initial state, and the replay is what is saved. Since the simulation is deterministic, the replay reproduces the executed result, which serves as a repeatability check. A trajectory is accepted if and only if its final state satisfies every predicate in $G$ and it passes the checks on interpenetration, unexpected contact, posture, smoothness, and joint limits. Rejected trajectories carry a stage code and a cause code, which the agent uses to revise the program. The video reference and the program are produced once per task; afterwards, trajectories are generated without API calls or human intervention, so the authoring cost is amortized over all trajectories of the task (\cref{sec:s_tasks}).

\section{Dataset Construction Details}
\label{sec:s_data}
This section complements Sec.~5 of the main paper. \Cref{tab:s_data} summarizes the construction funnel and the resulting dataset.

\begin{table}[t]
\centering
\caption{RoboHome dataset construction summary (Sec.~5).}
\label{tab:s_data}
\small
\setlength{\tabcolsep}{4pt}
\begin{tabular}{@{}lr@{}}
\toprule
Item & Count \\
\midrule
\multicolumn{2}{@{}l}{\emph{Floorplans (RPLAN~\cite{RPLAN})}} \\
\quad Source floorplans & 80{,}788 \\
\quad Valid floorplans retained & 44{,}036 \\
\quad Shell spot-check pass rate & 596/600 \\
\midrule
\multicolumn{2}{@{}l}{\emph{Assets}} \\
\quad Multi-source assets & 120K+ \\
\quad Materials & 3{,}728 \\
\quad Usable models after repair and pre-screening & 23{,}221 \\
\quad Runtime-qualified assets & 19{,}795 \\
\midrule
\multicolumn{2}{@{}l}{\emph{Generated data}} \\
\quad Whole-home interactive scenes & 20K+ \\
\quad Long-horizon tasks / embodiments & 6 / 3 \\
\quad Trajectories (duration) & 60K+ (${\approx}$750\,h) \\
\bottomrule
\end{tabular}
\end{table}

\noindent\textbf{Sources.}
Floorplans come from RPLAN~\cite{RPLAN}. Assets come from HSSD~\cite{HSSD}, Objaverse~1.0~\cite{Objaverse}, ArtVIP~\cite{ArtVIP}, ReplicaCAD~\cite{Habitat2}, and YCB~\cite{YCB}, together with 3{,}728 materials. Each asset admitted to the expert-annotated set $\mathcal{A}$ is qualified on eight annotated attributes, including category, style, appearance, original size, functional front, and mounting conditions (Sec.~3.1); these attributes drive retrieval, the size constraints of \cref{eq:s_size}, wall and support placement, and the orientation terms of \cref{tab:s_soft}.

\noindent\textbf{Floorplan and asset filtering.}
Of the 80{,}788 source floorplans, 44{,}036 valid floorplans are retained after normalization into building shells, and a spot check of 600 generated shells passes 596. Assets are repaired and pre-screened, which yields 23{,}221 usable models; of these, 19{,}795 pass the runtime qualification and are retained. Assets are always used at their original dimensions.

\noindent\textbf{Scene construction.}
The scene library is built with RoboHome so that every scene can be loaded directly for robot data generation and policy training. For each scene, a building shell is generated from a retained floorplan (\cref{sec:s_inputs}), compatible assets are retrieved and assembled at their original scale (\cref{sec:s_envelope,sec:s_joint}), and an end-to-end mapping among floorplan, shell, requirements, and object instances is kept in the context manifest. Scenes are exported to Isaac Sim~\cite{IsaacSim}. Post-save re-checks remove scenes with interpenetration, floating objects, or blocked doors, and an articulated part is marked as interactive only if it passes the readback test of opening and joint motion (\cref{sec:s_output}).

\noindent\textbf{Trajectory construction.}
Using the context manifests and the pipeline of \cref{sec:s_traj}, we generate trajectories for the Unitree G1, the RoboParty RP1, and Luna on six long-horizon loco-manipulation tasks: fruit in basket, box carrying, trash disposal, fridge placement, switch pressing, and drawer storage (\cref{sec:s_tasks}). Every trajectory in the dataset has passed the acceptance checks under deterministic open-loop replay (\cref{sec:s_traj}). The quantitative generation and policy results in the main paper use the Unitree G1; RP1 and Luna are included to demonstrate transfer of the pipeline to other embodiments.

\noindent\textbf{Hierarchical, physical, and motion annotation.}
Following 3D scene graphs~\cite{3DSceneGraph}, all levels share one coordinate system, scale, and orientation convention.
\emph{Scene level:} room topology, door connectivity, entrance anchors, and an omission ledger that records requested items that could not be placed.
\emph{Instance and physics level}~\cite{Habitat2}: categories, 6-DoF poses, support and collision bodies, mass and contact parameters, and the joint axes, limits, and open/closed states of doors, cabinets, and drawers.
\emph{Trajectory level:} the 50\,Hz whole-body reference sequence (40-dimensional for G1), ground-truth object states, multi-view renderings, and task predicates, all time-aligned.

\noindent\textbf{Scene information quality.}
Physical fields are tagged with their provenance through a scene information quality (SIQ) label that distinguishes measured, natively declared, estimated, and default parameters, so that users can tell which physical properties come from the source assets and which were filled in.

\noindent\textbf{Splits and release.}
Failure records are retained alongside accepted data. Training and test sets are separated by source floorplans and by held-out seeds. The dataset will be released after licensing review.

\section{Evaluation Protocol and Metric Definitions}
\label{sec:s_metrics}
This section complements Sec.~6.1 of the main paper with the full metric definitions and thresholds, the success criteria, the configuration of each method, and the scope of the timing. Throughout, \emph{RoboHome} denotes the framework, the method under evaluation, the released scene library, and the trajectory dataset.

\subsection{Test Inputs and Attempts}
\label{sec:s_inputs_eval}
We use 20 inputs based on distinct RPLAN floorplans. Each input comprises a scene description (a text requirement, a structured object list, and room functions) and a floorplan image. The explicit requirements of every input are frozen before any method is run. All methods receive the same inputs; a baseline that cannot accept a modality runs without it (\eg, without the floorplan image) and uses its native asset pool. Each method runs the 20 inputs with three initial seeds $s\in\{42,123,2026\}$, input $k\in\{0,\dots,19\}$ using seed $s{+}k$ for every method, which gives $N{=}60$ attempts per method. Every attempt has the same time budget $T_{\max}{=}90$\,min; an attempt that exceeds it is terminated and counted as a failure.

\subsection{Scene Quality Metrics}
\label{sec:s_quality}
The eight objective metrics of Tab.~3 measure the properties a home needs before a robot can use it for loco-manipulation: the task and household objects are present ($C_{\mathrm{req}}$, $C_{\mathrm{func}}$), furniture is arranged as it is used ($R_{\mathrm{rel}}$), the robot can walk from the entrance to every room and stand where objects are used ($R_{\mathrm{room}}$, $R_{\mathrm{use}}$), and the geometry is valid for physics simulation ($P_{\mathrm{col}}$, $P_{\mathrm{out}}$, $S_{\mathrm{sup}}$). They are computed by a generator-independent evaluator adapted from SceneEval~\cite{SceneEval} and PhyScene~\cite{PhyScene}, which is separate from the acceptance checks used during generation (\cref{sec:s_accept}). Each metric is computed per saved scene as a percentage and averaged over the scenes a method saved; attempts without a saved scene do not enter these averages but are counted in Yield and SR (\cref{sec:s_gen}). \Cref{tab:s_metric_defs} gives the definitions and thresholds. A dimension score in Tab.~3 is the mean of its metrics after replacing each lower-is-better metric $P$ by $100-P$; \eg, geometric validity is $\big[(100-P_{\mathrm{col}})+(100-P_{\mathrm{out}})+S_{\mathrm{sup}}\big]/3$.

\begin{table*}[t]
\centering
\caption{Definitions and thresholds of the eight objective scene-quality metrics (Tab.~3), as implemented in the independent evaluator. Each metric is a percentage computed per saved scene and averaged over the saved scenes of a method.}
\label{tab:s_metric_defs}
\small
\setlength{\tabcolsep}{4pt}
\begin{tabularx}{\linewidth}{@{}llcY@{}}
\toprule
Dimension & Metric & Dir. & Definition and thresholds (per scene) \\
\midrule
\makecell[lt]{Functional\\completeness} & Requirement coverage $C_{\mathrm{req}}$ & $\uparrow$ & Fraction of the entries of the frozen requirement list of the input that are matched one-to-one by an object whose asset-label category is the required one and whose projected center lies in the required room. \\
& Function coverage $C_{\mathrm{func}}$ & $\uparrow$ & Fraction of the basic furniture required by the function of each room (\eg, a bed in a bedroom, a toilet in a bathroom) that is present in that room, under the same matching rule. \\
\midrule
Layout plausibility & Relation satisfaction $R_{\mathrm{rel}}$ & $\uparrow$ & Fraction of the evaluated functional relations that hold, \eg, nightstand beside bed, chairs facing the dining table, TV facing the sofa. \\
\midrule
Usability & Room reachability $R_{\mathrm{room}}$ & $\uparrow$ & Fraction of essential rooms reachable from the entrance by a disc of radius $r{=}0.175$\,m, computed on a 2\,cm occupancy grid over the 0.05--1.8\,m height band with 8-connectivity. A room counts as reached if the reachable region overlaps its core region (the room's bounding box shrunk by 0.3\,m and intersected with the room) by more than 0.25\,m$^2$. \\
 & Usage-zone reachability $R_{\mathrm{use}}$ & $\uparrow$ & Fraction of furniture usage zones that can host a robot stance: at most 5\% of the zone is blocked and it overlaps the region reachable from the entrance by at least 0.01\,m$^2$. Movable chairs and stools, soft objects, and rugs are not treated as fixed obstacles. Room-to-room passages alone do not satisfy it. \\
\midrule
\makecell[lt]{Geometric\\validity} & Collision rate $P_{\mathrm{col}}$ & $\downarrow$ & Fraction of objects that interpenetrate another object or the shell. For each pair of exported visible meshes, the penetration depth is the largest depth by which a triangle of one passes the plane of an intersecting triangle of the other; a pair collides if this depth exceeds 10\,$\mu$m. Legal contacts are excluded by a category-pair list. \\
 & Out-of-bound rate $P_{\mathrm{out}}$ & $\downarrow$ & Fraction of objects that extend horizontally beyond their room, below the floor, or above the ceiling by more than 1\,cm. \\
 & Valid support $S_{\mathrm{sup}}$ & $\uparrow$ & Fraction of objects for which at least one installation mode allowed for their category holds: floor, surface, or ceiling with a gap of at most 1\,cm, or wall with a gap of 0--10\,mm. Surface support is tested at $5{\times}5$ footprint samples and requires the supporting object to lie under the object's center. \\
\bottomrule
\end{tabularx}
\end{table*}

\noindent\textbf{Missing evidence.}
Holodeck uses native AI2-THOR assets~\cite{AI2-THOR}, some of which lack collision or support evidence. For Holodeck, $P_{\mathrm{col}}$ is therefore computed only on the checkable subset of objects, and objects without evidence count as invalid in $S_{\mathrm{sup}}$ (marked $^\ast$ in Tab.~3).

\subsection{Generation Performance}
\label{sec:s_gen}
\noindent\textbf{Yield, SR, and time.}
Yield is the fraction of all $N$ attempts that save a simulator-loadable scene within $T_{\max}$; it does not require any validation check to pass. SR is the fraction of attempts that, within $T_{\max}$, save a loadable scene that passes every acceptance condition in \cref{tab:s_success}:
\begin{equation}
\mathrm{SR}=\frac{100\%}{N}\sum_{i=1}^{N}u_i,
\label{eq:s_sr}
\end{equation}
where $u_i{=}1$ if attempt $i$ saves such a scene and $u_i{=}0$ otherwise. Both rates use all $N$ attempts as the denominator. $T_{\mathrm{gen},i}=t_i^{\mathrm{end}}-t_i^{\mathrm{start}}$ runs from reading the input to saving the final scene.

\noindent\textbf{Success criteria.}
No aggregate score threshold is used: SR does not depend on any value of Tab.~3 reaching a preset percentage. An attempt succeeds only if all conditions in \cref{tab:s_success} hold, and a condition that lacks evidence or cannot be determined counts as not satisfied. Failures are of two kinds. A \emph{generation failure} saves no readable scene within the budget (timeout, abnormal process exit, or unreadable output); such attempts lower Yield and SR and are excluded from the quality and perception metrics. A \emph{non-compliant scene} is saved but fails at least one condition; it counts toward Yield but not SR.

\begin{table}[t]
\centering
\caption{Acceptance conditions for SR, checked by RoboHome on the saved scene. All conditions must hold.}
\label{tab:s_success}
\small
\begin{tabular}{@{}l>{\raggedright\arraybackslash}p{0.70\columnwidth}@{}}
\toprule
Condition & Criterion \\
\midrule
Valid delivery & The scene can be reloaded and rendering finishes normally; the saved files are unmodified, the evidence is bound to them by hash, and no mandatory check is skipped \\
Geometry & Every saved physical mesh is re-checked. For each intersecting pair, the equivalent intersection depth $d{=}2V/S$ ($V$: intersection volume, $S$: its surface area; $V{\le}10^{-15}$\,m$^3$ counts as zero) must not exceed 10\,$\mu$m; contacts certified by evidence, such as furniture on a thin rug, are recorded separately. No object lies outside its room \\
Support & Each object rests on its recorded support parent with its recorded support type, within a contact tolerance of 10\,$\mu$m; wall-mounted objects sit 0--10\,mm from the wall; support by a rug requires derived contact evidence \\
Structure & The reloaded shell, doorways, and dimensions match the input \\
Traversal & With a disc of radius $r{=}0.175$\,m, the entrance, all required rooms, and all room doors are mutually reachable \\
Required items & All protected explicit entities of the generator's execution request are placed, plus bed, sofa, dining table, toilet, shower, basin, fridge, stove, and kitchen counter \\
Hard layout & The entrance is not blocked, rooms are mutually connected, door swings are unobstructed, and the stove and sink sit on the counter \\
Budget & Generation and saving finish within $T_{\max}{=}90$\,min, including the post-save re-check \\
\bottomrule
\end{tabular}
\end{table}

\noindent\textbf{Acceptance checks \vs independent scoring.}
\label{sec:s_accept}
Scene acceptance and objective scoring are separate evaluations. Acceptance (\cref{tab:s_success}) is a gate that the generator applies to its own saved scene; Tab.~3 reports post-hoc scores from a generator-independent evaluator over all saved scenes. The two use different measurements and reference data (\cref{tab:s_accept_vs_score}), so a scene that passes acceptance need not obtain perfect independent scores, and the values of $P_{\mathrm{col}}$, $C_{\mathrm{req}}$, and $S_{\mathrm{sup}}$ in Tab.~3 are not bounds implied by SR. Two cases illustrate the difference.
(i) \emph{Requirements.} Acceptance checks the entities of the generator's execution request, whereas $C_{\mathrm{req}}$ matches the frozen benchmark list by asset category and geometric room. When the benchmark list asks for a plant on the balcony and the execution request places greenery in the living room, acceptance holds but $C_{\mathrm{req}}$ misses one entry; this single entry is unmatched in 46 of the 55 scored RoboHome scenes.
(ii) \emph{Support.} A coffee table resting on a rug about 2\,cm above the floor is valid under acceptance, which follows the recorded support parent (the rug) with contact evidence, but invalid under $S_{\mathrm{sup}}$, whose category rule allows a coffee table only on the floor within 1\,cm.
These differences explain non-perfect independent scores; they do not relax acceptance, and SR counts only attempts whose saved scene passes all conditions of \cref{tab:s_success}.

\begin{table*}[t]
\centering
\caption{Differences between the generation-time acceptance checks (used for SR, \cref{tab:s_success}) and the independent evaluator (used for Tab.~3, \cref{tab:s_metric_defs}).}
\label{tab:s_accept_vs_score}
\small
\setlength{\tabcolsep}{4pt}
\begin{tabular}{@{}l>{\raggedright\arraybackslash}p{0.40\textwidth}>{\raggedright\arraybackslash}p{0.44\textwidth}@{}}
\toprule
Aspect & Acceptance (SR) & Independent scoring (Tab.~3) \\
\midrule
Collision & Saved physical geometry; equivalent solid-intersection depth $2V/S$ ${>}10\,\mu$m is illegal; evidence-certified contacts allowed & Exported visible meshes; triangle-level plane-penetration depth ${>}10\,\mu$m counts as a collision; legal contacts from a category-pair list \\
Support & Recorded support parent and support type; 10\,$\mu$m contact tolerance; wall 0--10\,mm; rug support with evidence & Category-based installation modes; floor, surface, or ceiling gap ${\le}1$\,cm; wall 0--10\,mm \\
Requirements & Generator's execution request, its protected explicit entities, and the omission ledger & Frozen benchmark list; one-to-one match by asset category and the room containing the projected center \\
Access & Disc of radius 0.175\,m connects the entrance, all rooms, and all room doors & $R_{\mathrm{room}}$: overlap with room core regions on a 2\,cm grid; $R_{\mathrm{use}}$: clearance and reachability of usage zones \\
\bottomrule
\end{tabular}
\end{table*}

\noindent\textbf{Traversal radius.}
The radius $r{=}0.175$\,m equals the walking radius $r_b$ that RoboHome uses during generation (\cref{sec:s_inputs}). It is a geometric parameter of the check rather than the footprint of a particular robot: it is below the standing footprints of the Unitree G1, the RP1, and Luna. A scene that fails the check therefore cannot be traversed by any of them, whereas passing it is necessary but not sufficient for traversal by a given embodiment; embodiment-specific feasibility is established by trajectory generation (\cref{sec:s_traj}).

\subsection{Visual and Human Evaluation}
\label{sec:s_visual}
\noindent\textbf{Judge models.}
GPT-6.1 Sol, GPT-6 Astra, and Claude Opus 5.5 rate visual aesthetics and visual realism on a 1--5 scale. All generated scenes (Holodeck 38, EmbodiedGen V2 29, RoboHome 55) are rendered identically and anonymized, so the judges do not know which method produced a scene, and each judge receives the same renderings and requirement summaries. Each judge scores every scene in two independent rounds; we report the two-round mean per judge (\cref{tab:s_judges}) and the mean over the three judges (Tab.~4).

\noindent\textbf{Agreement.}
Kendall's $W$ measures the agreement of the three judges on the ranking of the three methods; $W{=}1.00$ means that all judges produce the same ranking on both dimensions. ICC$(2,k)$ (two-way random effects, average of $k{=}3$ raters) measures the scene-level reliability of the three-judge mean; it is 0.75 for aesthetics and 0.63 for realism.

\noindent\textbf{Cross-validation.}
Because model ratings may deviate systematically from human judgment, the judge panel is cross-checked in two ways. Q-Align~\cite{QAlign}, a pretrained scorer trained on human-aligned data, rates image quality and aesthetics on its own 1--5 scale (\cref{tab:s_qalign}); its scores are an auxiliary check and are not interchangeable with the judge-panel scores. In the human study, 59 volunteers, including embodied AI practitioners, view the same anonymized renderings in random order and rate layout plausibility, aesthetics, and realism on the same 1--5 scale. We report the mean and the standard deviation across scenes for each method.

\subsection{Methods and Implementation}
\label{sec:s_impl}
\noindent\textbf{Baselines.}
The language model called inside Holodeck~\cite{Holodeck} and SceneSmith~\cite{SceneSmith} is unified to GPT-5.6 Sol. EmbodiedGen V2~\cite{EmbodiedGen-V2} uses its native text-input pipeline with the Infinigen complexity set to medium. Each baseline uses its native asset pool. SceneSmith saved no scene within the budget in any of the 60 attempts and is therefore omitted from Tabs.~3 and~4.

\noindent\textbf{RoboHome.}
RoboHome calls no commercial model for scene generation. The floorplan image is parsed by Qwen3-VL~\cite{Qwen3-VL} run locally, and the object list is built by rules from the structured object list and the room functions (\cref{sec:s_inputs}). The search configuration is given in \cref{sec:s_joint}.

\noindent\textbf{Timing scope and hardware.}
$T_{\mathrm{gen}}$ covers everything from reading the input to saving the final scene; for RoboHome this includes shell construction and the post-save re-check. Shared asset preprocessing, rendering, and offline evaluation are not counted. RoboHome runs on CPU only, on a server with 192 logical cores (Intel Xeon Platinum 8558), in 12 parallel task streams, each with its own copy of the asset cache. The baselines run on an Intel Xeon Gold 6530 host with 8 logical cores per process. Because the hosts differ, $T_{\mathrm{gen}}$ is not a like-for-like speed comparison across methods.

\section{Supplementary Visual Evaluation}
\label{sec:s_judges}
This section complements Tab.~4 of the main paper. All three evaluations below score the same generated scenes (Holodeck 38, EmbodiedGen V2 29, RoboHome 55) from the same anonymized renderings (\cref{sec:s_visual}).

\noindent\textbf{Per-judge scores.}
\Cref{tab:s_judges} lists the scores of each judge model behind the model mean of Tab.~4, together with the human ratings. All three judges rank RoboHome first and Holodeck second on both dimensions. The human ratings show a smaller aesthetic margin over Holodeck (2.65 \vs 2.56) than the judges do, which is why both are reported.

\begin{table*}[t]
\centering
\caption{\textbf{Per-judge subjective visual scores} (1--5 scale, $\uparrow$). Three judge models blindly scored all generated scenes (Holodeck 38, EmbodiedGen V2 29, RoboHome 55) in two rounds; each judge row reports its two-round mean. Subscripts: standard deviation across the three judges for ``Model mean''; standard deviation across scenes for ``Human'' (59 volunteers). Kendall's $W{=}1.00$ across judges on both dimensions; scene-level ICC$(2,k)$ of the model mean: 0.75 (aesthetics) and 0.63 (realism). \textbf{Bold}/\underline{underline}: best/second best per row within each dimension.}
\label{tab:s_judges}
\small
\setlength{\tabcolsep}{4.5pt}
\fitlinewidth{%
\begin{tabular}{@{}lcccccc@{}}
\toprule
& \multicolumn{3}{c}{Visual aesthetics} & \multicolumn{3}{c}{Visual realism} \\
\cmidrule(lr){2-4}\cmidrule(lr){5-7}
Rater & Holodeck~\cite{Holodeck} & \makecell{EmbodiedGen\\V2~\cite{EmbodiedGen-V2}} & \textbf{RoboHome} & Holodeck~\cite{Holodeck} & \makecell{EmbodiedGen\\V2~\cite{EmbodiedGen-V2}} & \textbf{RoboHome} \\
\midrule
GPT-6.1 Sol & \underline{2.66} & 2.34 & \textbf{2.92} & \underline{2.50} & 2.24 & \textbf{2.84} \\
GPT-6 Astra & \underline{2.41} & 2.07 & \textbf{2.86} & \underline{2.33} & 2.24 & \textbf{2.58} \\
Claude Opus 5.5 & \underline{2.21} & 2.00 & \textbf{2.60} & \underline{2.22} & 2.00 & \textbf{2.38} \\
\midrule
Model mean & \underline{2.43}$_{\pm0.22}$ & 2.14$_{\pm0.18}$ & \textbf{2.79}$_{\pm0.17}$ & \underline{2.35}$_{\pm0.14}$ & 2.16$_{\pm0.14}$ & \textbf{2.60}$_{\pm0.23}$ \\
Human & \underline{2.56}$_{\pm0.48}$ & 2.14$_{\pm0.35}$ & \textbf{2.65}$_{\pm0.48}$ & \underline{2.26}$_{\pm0.45}$ & 2.03$_{\pm0.19}$ & \textbf{2.53}$_{\pm0.54}$ \\
\bottomrule
\end{tabular}}
\end{table*}

\noindent\textbf{Q-Align.}
\Cref{tab:s_qalign} reports the Q-Align~\cite{QAlign} scores. Q-Align gives RoboHome the highest aesthetic score (3.13), above Holodeck (2.92) and EmbodiedGen V2 (2.45). Its quality score for RoboHome (3.97) is slightly below Holodeck (4.05) and above EmbodiedGen V2 (3.67). This independent pretrained scorer therefore supports the visual-aesthetic advantage of RoboHome found by the judge panel. Q-Align measures image quality and aesthetics only; it does not measure floorplan fidelity or whether a robot can use the scene, which Tab.~3 and the human layout ratings cover. Taken together with the judge-panel and human evaluations, these results indicate that RoboHome improves the scene organization needed for robotic loco-manipulation while keeping visual quality and aesthetics comparable to or better than the baselines.

\begin{table}[t]
\centering
\caption{\textbf{Q-Align scores} of the generated whole-home scenes (1--5 scale, $\uparrow$), averaged over the scenes generated by each method. This pretrained scorer is an auxiliary cross-check of the judge-panel and human evaluations; its scale is not interchangeable with Tab.~4. \textbf{Bold}/\underline{underline}: best/second best per column.}
\label{tab:s_qalign}
\small
\begin{tabular}{@{}lccc@{}}
\toprule
Method & \#Scenes & Quality $\uparrow$ & Aesthetics $\uparrow$ \\
\midrule
Holodeck~\cite{Holodeck} & 38 & \textbf{4.05} & \underline{2.92} \\
EmbodiedGen V2~\cite{EmbodiedGen-V2} & 29 & 3.67 & 2.45 \\
\textbf{RoboHome} & 55 & \underline{3.97} & \textbf{3.13} \\
\bottomrule
\end{tabular}
\end{table}

\section{Ablation Configurations and Qualitative Results}
\label{sec:s_ablation}
This section complements Sec.~6.3 and Tab.~5 of the main paper. Each variant removes one component of RoboHome, and the comparison asks which component makes the generated homes usable for robot tasks.

\noindent\textbf{Protocol.}
Each variant runs the same 20 inputs as the main comparison with initial seed $s{=}42$ (seed $42{+}k$ for input $k$), \ie, 20 attempts per configuration. Inputs, candidate assets, evaluators, and time budgets are fixed; only the ablated component changes. Yield and SR use all 20 attempts as the denominator; the other metrics of Tab.~5 are computed on the scenes saved within the budget, and $T_{\mathrm{gen}}$ is the median time per scene. The ``w/o expert-annotated assets'' variant ran on a different host, so its $T_{\mathrm{gen}}$ is not comparable.

\noindent\textbf{Variants.}
\emph{w/o envelope pre-layout:} the envelope stage (\cref{sec:s_envelope}) is removed and the joint asset--pose search (\cref{sec:s_joint}) runs directly.
\emph{w/o constraint metrics:} the spatial and functional constraints of the pre-layout and the cost $\ell$ (Eq.~(3)) are removed, keeping only boundary and collision checks.
\emph{w/o expert-annotated assets:} retrieved assets bypass expert annotation and qualification (\cref{sec:s_data}).

\begin{figure*}[t]
  \centering
  \includegraphics[width=0.95\textwidth]{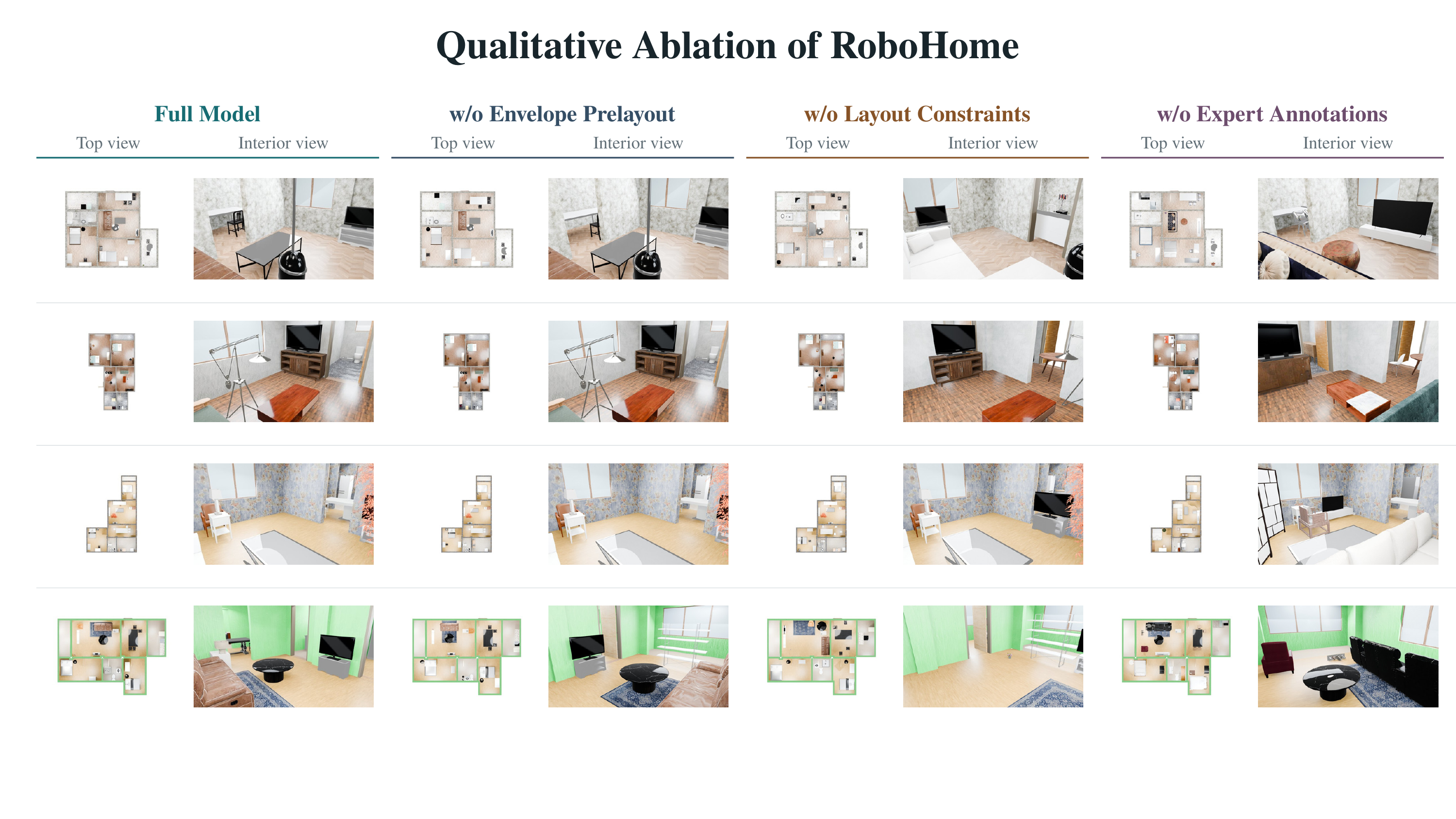}
  \caption{\textbf{Qualitative ablation} on four inputs shared across all configurations (rows). Each cell shows a top view (left) and a selected interior view (right). Columns, left to right: full method; w/o envelope pre-layout; w/o constraint metrics (``Layout Constraints'' in the artwork); w/o expert-annotated assets (``Expert Annotations'' in the artwork). ``RoboHome'' denotes our scene generation method.}
  \label{fig:s_ablation}
\end{figure*}

\noindent\textbf{Qualitative results.}
\Cref{fig:s_ablation} compares the four configurations on four inputs shared by all of them. Without envelope pre-layout (second column), the scenes that are produced keep similar furniture arrangements and differ mostly locally, consistent with Tab.~5: pre-layout mainly raises SR (95\% \vs 40\%) rather than the quality of the layouts that are produced. Without constraint metrics (third column), spatial relations become implausible; for example, a TV blocks a doorway passage, which would cut the robot's route between rooms. This matches the drop in layout plausibility (95.4\% to 42.8\%) and usability (87.8\% to 26.6\%) while geometric validity stays at 99.0\%: a collision-free layout is not necessarily one a robot can move through. Without expert-annotated assets (fourth column), tall cabinets and blocky assets obstruct passages that a robot would need, and the sofa, TV, and partition take unnatural relative orientations, matching the lower layout plausibility (87.5\%) and usability (78.5\%).

\section{Long-Horizon Tasks and Data Generation Details}
\label{sec:s_tasks}
This section complements Sec.~6.4, Fig.~5, and Tab.~6 of the main paper.

\noindent\textbf{Task specifications.}
\Cref{tab:s_tasks} specifies the six long-horizon loco-manipulation tasks executed by the Unitree G1 in Fig.~5. The keyframes of Fig.~5 are rendered in Isaac Sim by playing back the generated trajectories, which are simulated in MuJoCo. Together they cover countertop and floor grasping, bimanual carrying, pressing a fixed object, and interaction with articulated doors and drawers. Each task couples walking with manipulation, so the scene must provide compatible objects, interaction space, and passages throughout the sequence rather than only at an isolated manipulation pose. Box carrying explicitly traverses a doorway from the hallway into the bedroom; the other tasks involve single-room manipulation with local navigation. The quantitative results in Tab.~6 are for the Unitree G1; RP1 and Luna are included to demonstrate embodiment transfer.

\begin{table*}[t]
\centering
\caption{The six long-horizon loco-manipulation tasks of Sec.~6.4 (Fig.~5 and Tab.~6).}
\label{tab:s_tasks}
\small
\setlength{\tabcolsep}{5pt}
\begin{tabularx}{\linewidth}{@{}lXl@{}}
\toprule
Task & Description & Interaction \\
\midrule
Fruit in basket & Walk to the fruit, pick it up (\eg, a pear), and put it into the basket. & Countertop grasping, placing \\
Box carrying & Pick up a box from the hallway floor, carry it through a doorway into the bedroom, and place it on a low table. & Floor grasping, bimanual carrying \\
Trash disposal & Pick up a bottle and drop it into the bin. & Grasping, releasing into a container \\
Fridge placement & Open the fridge door and place a bottle inside. & Articulated door, placing \\
Switch pressing & Approach a wall switch and press it. & Pressing a fixed object \\
Drawer storage & Open a drawer, insert an item taken from the countertop, and close the drawer. & Articulated drawer, placing \\
\bottomrule
\end{tabularx}
\end{table*}

\noindent\textbf{Generation metrics.}
All three metrics are reported per task in Tab.~6.
\emph{Pass rate} $Y$ is the fraction of held-out seeds whose trajectories pass all acceptance checks (\cref{sec:s_traj}).
\emph{Per-trajectory cost} $C$ is the program-authoring API cost of the task, \ie, the coding agent's token usage for writing and refining the expert program priced at public Claude Opus~5.5 rates, divided by the 50 accepted trajectories of the task.
\emph{Generation time} $L$ is the total single-process generation time of the task, including failed attempts, divided by the number of passing trajectories.

\noindent\textbf{Aggregates.}
Across the six tasks of Tab.~6, $Y$ averages 87.3\% (range 67\%--100\%), $C$ averages \$0.92, and $L$ averages 134\,s; these are unweighted means over tasks. The per-hour cost in the main paper divides the total program-authoring API cost of the six tasks, $\sum_{\text{task}} 50\,C_{\text{task}}$, by the duration of the resulting 300 fine-tuning trajectories (approximately 3.75\,h, \ie, about 45\,s per trajectory), which gives about \$70 per hour of trajectory data. The values of $C$ in Tab.~6 are rounded to \$0.1.

\noindent\textbf{Cost scope.}
The API cost covers only synthesizing and refining each expert program. It excludes video generation, simulation compute, and scene generation. Once a program is finalized, further trajectories are generated in parallel processes without API calls or human intervention, so this one-off cost is spread over all additional accepted trajectories and the per-hour figure decreases as more data are generated.

\noindent\textbf{Efficiency by task.}
Efficiency varies with task length. Switch pressing, fruit in basket, and trash disposal have fewer stages and shorter trajectories: $Y\ge96\%$, $L$ of 50--75\,s, and $C\le\$0.5$. Box carrying, drawer storage, and fridge placement have more stages and longer trajectories: their pass rates are 86\%, 75\%, and 67\%, with $L$ of 154--246\,s and $C$ of \$1.3--\$1.6. For these three tasks, failed attempts account for 17--26\% of the generation time.

\noindent\textbf{Policy evaluation protocol.}
For each task, 50 passing trajectories from the generated batch are used to fine-tune ACT~\cite{ACT}, $\pi_{0.5}$~\cite{pi05}, and GR00T N1.7~\cite{gr00t}. All three policies output the same 40-dimensional whole-body reference action as the expert trajectories, which is executed by the same frozen tracking policy. Closed-loop evaluation uses 20 trials per task with seeds unseen during training and the same judge as trajectory generation, so each success rate in Tab.~6 is a multiple of 5\%.

\noindent\textbf{Generation pass rate \vs policy success.}
The two stages measure different things. Box carrying has an 86\% generation pass rate but at most 35\% policy success, whereas trash disposal reaches 100\% and 75\%. Reliably generating expert trajectories therefore does not remove the difficulty of learning the complete task from 50 demonstrations; tasks that involve carrying or articulated interaction (box carrying, fridge placement, drawer storage) remain the hardest for all three policies.

\end{document}